\documentclass[11pt]{article}

\usepackage[preprint]{acl}

\usepackage{times}
\usepackage{latexsym}

\usepackage[T1]{fontenc}
\usepackage[utf8]{inputenc}

\usepackage{microtype}
\usepackage{inconsolata}
\usepackage{subcaption}
\usepackage{graphicx}
\usepackage{booktabs}
\usepackage{amsmath}
\usepackage{amssymb}

\usepackage{graphicx}
\graphicspath{{figures/}}

\usepackage{url}
\usepackage[size=tiny]{todonotes}

\title{ValueGround: Evaluating Culture-Conditioned \\ 
Visual Value Grounding in MLLMs
    
}

\author{
\textbf{Zhipin Wang$^{1}$, Christoph Leiter$^{1}$, Christian Frey, Mohamed Hesham Ibrahim Abdalla,} \\
\textbf{ Josif Grabocka, Steffen Eger$^{1}$}\\
University of Technology Nuremberg (UTN), Germany\\
\texttt{zhipin.wang@utn.de, steffen.eger@utn.de} \\
$^1$\texttt{NLLG} (\url{https://nl2g.github.io/})
}

\begin{document}
\maketitle

\begin{abstract}
Cultural values are expressed not only through language but also through visual scenes and everyday social practices. Yet existing evaluations of cultural values in language models are almost entirely text-only, leaving it unclear whether culture-conditioned judgments remain stable when response options are visualized. We introduce \textsc{ValueGround}, a benchmark for evaluating \emph{culture-conditioned visual value grounding} in multimodal large language models (MLLMs). Built from World Values Survey questions, \textsc{ValueGround} uses minimally contrastive image pairs to represent opposing response options while controlling irrelevant variation. Given a country, a question, and an image pair, a model must choose the image that best matches the country’s value tendency without access to the original response-option texts. Experiments across six MLLMs and 13 countries show that models perform substantially worse with visualized response options than with the original textual options, with average accuracy dropping from 72.8\% to 62.6\%. Our benchmark provides a controlled testbed for studying cross-modal transfer of culture-conditioned value judgments.
\end{abstract}

\section{Introduction}

\begin{figure}[ht]
    \centering
    \includegraphics[width=\columnwidth]{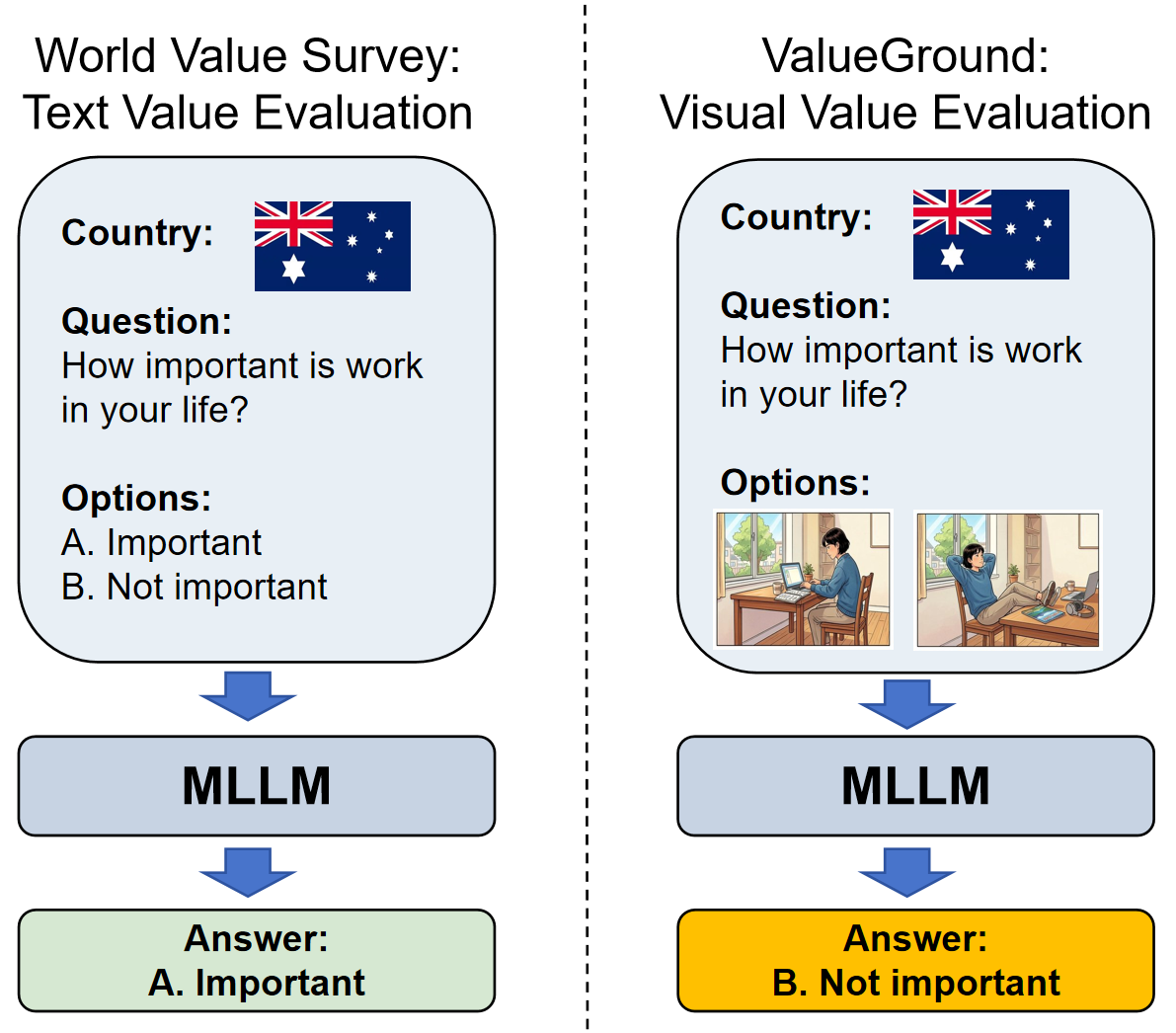}
    \caption{
    A motivating reversal example. For the same country-question pair, the model predicts one response when the answer choices are textual, but reverses its prediction when the response options are replaced with visual proxies. This example illustrates the central challenge in \textsc{ValueGround}: whether models can maintain consistent culture-conditioned judgments when response options are visualized.
    }
    \label{fig:reversal-example}
\end{figure}

\begin{figure*}[ht]
    \centering
    
    \includegraphics[width=\linewidth]{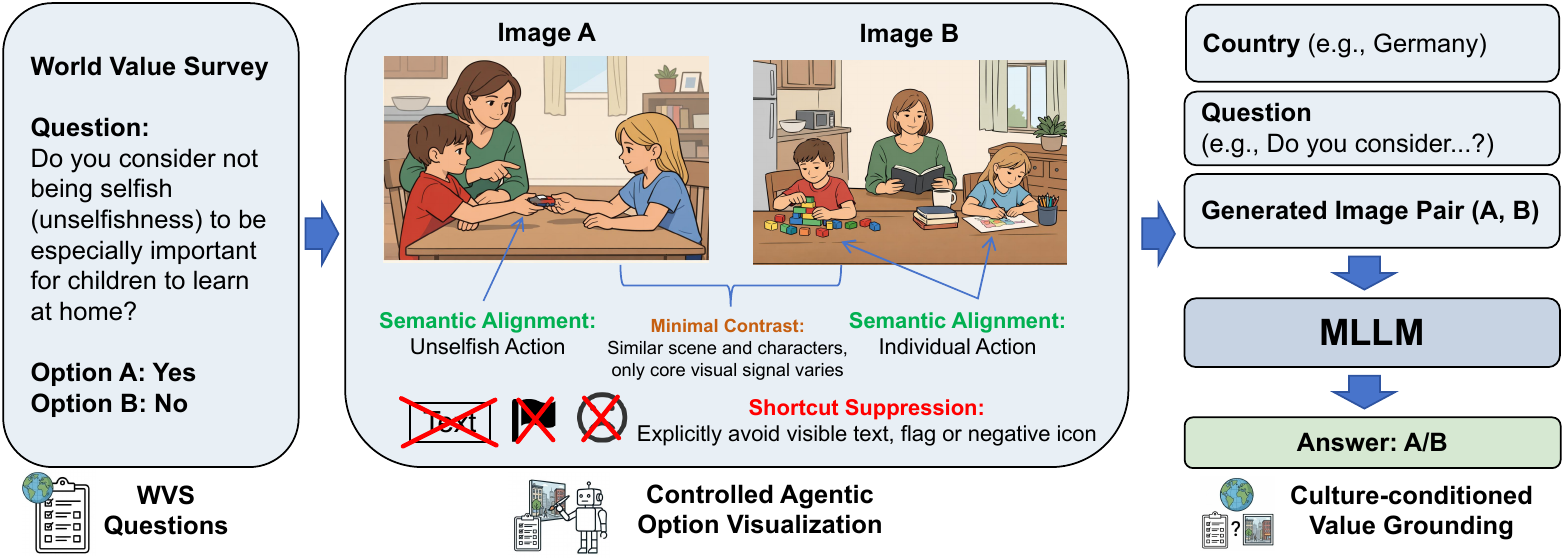}
    \caption{
    Overview of \textsc{ValueGround}. Starting from a value question $q$ with two response options $(o_A, o_B)$, we construct a controlled image pair $(I_A, I_B)$ that reflects the question and two option meanings while minimizing irrelevant differences and shortcut cues. In the main task, the model receives a country $c$, the question $q$, and the image pair $(I_A, I_B)$, and must choose which image better matches the value tendency associated with that country. The response-option texts are hidden at test time.
    }
    \label{fig:valueground-overview}
\end{figure*}

Cultural values like social behavior, public norms, and everyday choices shape judgments. 
Recent work has therefore begun to evaluate whether language models reflect culture-conditioned value tendencies, using cross-national survey resources such as the World Values Survey (WVS) \citep{wvs2022wave7} and related benchmarks  \citep{santurkar2023whose, zhao2024worldvaluesbench, durmus-etal-2024-globalopinionqa, alkhamissi-etal-2024-investigating}. However, this line of work remains almost entirely text-only: models are given a country, a survey question, and verbal response options. It remains unclear whether the same judgments survive when the response options are visualized.

Cultural values are expressed not only through language, but also through visual scenes and everyday social practices. From a human perspective, such values can also be learned through narrative fiction, picture books, and observation of everyday social situations, especially during childhood, in addition to abstract verbal definitions \citep{ bandura1977social, gasser2022children, chen2025effectiveness}.
As multimodal language models increasingly operate over both text and images 
\citep{mistral_small_2506,gemmateam2025gemma3technicalreport,qwen_qwen3vl,anthropic2025haiku45,singh2025openaigpt5card,google_gemini3flashpreview}, cultural assessment should also test whether value judgments remain stable when the same option meanings are conveyed visually rather than stated verbally. Inspired by prior work on visio-linguistic compositional evaluation \citep{thrush2022winoground} and documented language bias \citep{zhao-etal-2025-looking} in MLLMs, visualizing response options therefore provides a controlled way to test whether culture-conditioned judgments remain stable when the same semantic contrast is presented through images rather than text. As shown in Figure~\ref{fig:reversal-example}, the same model may answer one way when response options are verbalized, yet reverse its prediction when the same option meanings are presented visually. Such prediction reversals provide a direct test of whether MLLMs can preserve culture-conditioned judgments across modalities.

Motivated by the largely text-only nature of current evaluation, we introduce \textsc{ValueGround}\footnote{Project repository: \url{https://github.com/NL2G/ValueGround}}, a benchmark for evaluating culture-conditioned judgments when response options are conveyed \emph{visually} under controlled \emph{response-option substitution}. Survey response options often express abstract social tendencies, norms, or preferences rather than concepts with a single canonical visual form, making them difficult to visualize without introducing irrelevant cues. We therefore operationalize each option as a carefully constructed image within a minimally contrastive pair. As illustrated in Figure~\ref{fig:valueground-overview}, \textsc{ValueGround} is guided by three principles: \emph{semantic alignment}, so that each image reflects its target option meaning; \emph{minimal contrast}, so that the paired images share the same scene and differ primarily in the option-defining signal; and \emph{shortcut suppression}, so that visible text or other superficial cues do not trivially reveal the answer.

In the resulting task, the model is given a country $c$, a question $q$, and an image pair $(I_A, I_B)$, and must choose which image better matches the value tendency associated with that country, without access to the original response-option texts. This formulation lets us test whether models that succeed with verbal options can preserve the same country-conditioned judgments when those options are visualized.

Our contributions are:
\begin{itemize}
    \item We introduce \textsc{ValueGround}, the first controlled benchmark for testing whether culture-conditioned value judgments remain stable when textual response options are visualized as minimally contrastive images.

    \item We propose an agentic construction and validation pipeline that converts World Values Survey response options into human-validated image pairs, ensuring semantic alignment with the intended options, minimal variation beyond the intended value contrast, and suppression of explicit shortcut cues.

    \item We evaluate six MLLMs across 13 countries and identify a persistent text-to-vision grounding gap: accuracy drops from 72.8\% in the text-only setting to 62.6\% with visualized response options, and remains lower even when models correctly identify which response option each image depicts. This indicates that \textsc{ValueGround} requires both image--option recognition and the application of country-conditioned value knowledge to visual evidence.
\end{itemize}

\section{Related Work}
\paragraph{Cultural values and alignment in LLMs.}
A growing body of work studies culture-conditioned value tendencies in LLMs. Early work probed such tendencies using established sociological frameworks such as Hofstede's cultural dimensions \citep{arora-etal-2023-probing}. Subsequent survey-based evaluations examined how LLMs reflect opinions and values across demographic, cultural, and national contexts \citep{santurkar2023whose,durmus-etal-2024-globalopinionqa,zhao2024worldvaluesbench,alkhamissi-etal-2024-investigating, ju-etal-2025-benchmarking}. More recent work explores explicit cultural alignment and culture-aware adaptation or fine-tuning \citep{li2024culturellm, xu-etal-2025-self, abdalla2025zhyper}. Related multimodal evaluation also studies broader human-value or preference alignment in MLLMs \citep{shi2024ch3ef,xu-etal-2024-multiskill,zhao-etal-2025-omnialign}, but without culture-conditioned grounding. Our work builds on this literature, but moves beyond text-only evaluation to a multimodal setting in which the country and question remain textual while answer options are visualized.

\paragraph{Multimodal cultural evaluation.}
Multimodal cultural evaluation has expanded from geographically diverse visual evaluation, such as GeoDE \citep{ramaswamy-etal-2023-geode}, to multicultural vision-language reasoning, such as MaRVL \citep{liu-etal-2021-visually}.
Recent multimodal benchmarks further evaluate culturally specific concepts, practices, and grounding \citep{nayak-etal-2024-benchmarking, bhatia-etal-2024-local, vayani2025all}, and GIMMICK \citep{schneider-etal-2025-gimmick} further broadens this line with globally inclusive multimodal cultural knowledge benchmarking. 
The closest work to ours is Beyond Words \citep{yadav-etal-2025-beyond}, which probes cultural value sensitivity in multimodal models by using images as proxies for cultural context. In contrast, \textsc{ValueGround} keeps the country and question explicit in text and visualizes only the answer options, directly testing value grounding rather than culture inference from images.

\paragraph{Multi-image and contrastive evaluation.}
Our benchmark is also related to multimodal evaluation settings that require models to rely on subtle visual differences or contextual evidence. Winoground \citep{thrush2022winoground} uses contrastive image--text pairs to probe whether models can align fine-grained distinctions with the correct alternative. Prior work further shows that MLLMs often struggle with fine-grained visual perception \citep{fu-etal-2024-blink, tong-etal-2024-mmvp}. CODIS \citep{luo-etal-2024-codis} studies context-dependent visual comprehension, where textual context changes how an image should be interpreted. MuirBench \citep{wang-etal-2024-muirbench} and MIBench \citep{liu-etal-2024-mibench} extend evaluation to multi-image understanding across diverse visual relations and task settings. In contrast, \textsc{ValueGround} focuses on grounding survey-derived value alternatives under explicit country conditioning with minimally contrastive paired images.

\section{Benchmark Construction}
\label{sec:method}

We first formalize the task setup of \textsc{ValueGround} in Section~\ref{sec:setup}. We then present an agentic framework for constructing controlled paired visualizations from WVS-style questions in Section~\ref{sec:image_construction}, followed by the pair-quality rubric used to assess candidate image pairs in Section~\ref{sec:rubric}.

\subsection{Problem Setup}
\label{sec:setup}

We retain only attitudinal WVS questions and exclude demographic items as well as questions whose answers are primarily factual or objective. Let $\tilde{Q}$ denote the resulting question set. For each question $q \in \tilde{Q}$, let $R_q=(r_1,\dots,r_{n_q})$ be its original ordered response-option list.

For the multimodal task, we reduce each question to a binary option contrast pair
\begin{equation}
O_q=(o_A,o_B)=
\begin{cases}
(r_1,r_2), & n_q=2,\\
(r_1,r_{n_q}), & n_q>2.
\end{cases}
\end{equation}
This reduction preserves the main value opposition of the question while avoiding fine-grained ordinal distinctions that are difficult to visualize reliably.

From $O_q$, we generate a paired visualization $I_q=(I_A,I_B)$ and reuse the same pair for all countries. At test time, the model receives the country $c$, the question $q$, and the image pair $I_q$, but not the verbal option pair $O_q$, and predicts
\begin{equation}
\hat y_{c,q}
=
\mathrm{MLLM}(c,q,I_q),
\qquad
\hat y_{c,q}\in\{o_A,o_B\}.
\end{equation}
 The prediction is evaluated against a binarized country-level label $y_{c,q}\in\{o_A,o_B\}$ derived from the original WVS response distribution. Two auxiliary text-only and alignment-only control settings are introduced later in Section~\ref{sec:eval_setup}.

\subsection{Agentic Paired-Image Construction Framework}
\label{sec:image_construction}

Given a question $q$ and a binary option pair $O_q=(o_A,o_B)$, our goal is to construct a paired visualization $I_q=(I_A,I_B)$ that expresses the two option directions while controlling irrelevant variation. We operationalize this goal with three criteria.

\begin{itemize}
\item \textbf{Semantic alignment.}
For each direction $d\in\{A,B\}$, image $I_d$ should express $o_d$ under question $q$ through concrete scene evidence, such as actions, interactions, object configurations, or contextual relations.

\item \textbf{Minimal contrast.}
The pair $(I_A,I_B)$ should share the same base scene, participants, framing, and visual style, and differ mainly in the attributes that realize the intended contrast, with unrelated variation minimized.

\item \textbf{Shortcut suppression.}
Neither image should contain direct answer-bearing cues, such as readable text, labels, flags, UI elements, check marks, or cross symbols.
\end{itemize}

We implement these criteria with three coordinating MLLM agents and an image generator. The framework fixes a shared scene and realizes the option contrast through controlled edits rather than independent generation. Figure~\ref{fig:pipeline} in Appendix~\ref{sec:appendix:edit-framework} provides an overview and additional implementation details.

\paragraph{Planner Agent.}
The Planner Agent takes $(q,O_q)$ as input and produces a structured plan
\begin{equation}
p_q=(s_q,l_q,e_q,\rho_q)=\mathcal{A}_{\text{plan}}(q,O_q),
\end{equation}
where the value semantics of $q$ define a plausible shared base scene $s_q$ and the \emph{locked attributes} $l_q$ that should remain fixed across the pair, while the contrasting tendencies in $O_q$ define the \emph{editable attributes} $e_q$ that should differentiate the two images. The plan also records risk factors $\rho_q$, such as visible text or symbolic shortcuts.

\paragraph{Base Image Generator.}
Conditioned on the question-derived shared components $(s_q,l_q)$, the Base Image Generator produces a neutral base image that fixes the cast, layout, framing, and visual style of the scene, while leaving the option-defining difference unresolved:
\begin{equation}
I^{\text{base}}_q=\mathcal{G}_{\text{base}}(s_q,l_q).
\end{equation}

\paragraph{Editor Agent.}
Starting from the base image $I^{\text{base}}_q$ and the plan $p_q$, the Editor Agent applies two constrained edits, one for each option direction, while preserving the locked attributes of the shared scene. This concentrates variation on the intended semantic axis and keeps the pair visually comparable:
\begin{equation}
\tilde I_q=\mathcal{A}_{\text{edit}}(I^{\text{base}}_q,p_q),
\qquad
\tilde I_q=(\tilde I_A,\tilde I_B).
\end{equation}

\paragraph{Critic Agent.}
The Critic Agent evaluates the candidate pair $\tilde I_q$ with respect to the question $q$ and option pair $O_q$, and outputs a revision decision $r_q$. It checks whether the two images match their respective option directions, whether they remain aligned on non-contrastive content, and whether explicit shortcut cues are present:
\begin{equation}
r_q = \mathcal{A}_{\text{critic}}(q,O_q,\tilde I_q).
\end{equation}
The output $r_q$ is one of \textsc{Accept}, \textsc{ReviseEdits}, \textsc{Regenerate}, or \textsc{Replan}. Here, \textsc{ReviseEdits} keeps the current base image and revises only the editing step, \textsc{Regenerate} requests a new base image under the current plan, and \textsc{Replan} restarts the process from planning. The loop continues until a pair is accepted or a retry budget is exhausted.

Any image can only partially instantiate an abstract value tendency. The goal of the framework is therefore not exhaustive depiction, but a controlled paired visualization in which the intended option contrast is visually recoverable.

\subsection{Pair-Quality Rubric}
\label{sec:rubric}

To assess the quality of candidate image pairs, we use a four-item rubric aligned with the three construction criteria introduced in Section~\ref{sec:image_construction}. Specifically, \emph{semantic alignment} is decomposed into two directional checks: \textsc{Match-A} ($Q_1 \in \{0,1,2\}$), which evaluates whether image $I_A$ fits the question and option direction $o_A$, and \textsc{Match-B} ($Q_2 \in \{0,1,2\}$), which does the same for $I_B$ and $o_B$. For $Q_1$ and $Q_2$, a score of 2 indicates clear semantic alignment, 1 indicates that the intended option direction is visually distinguishable but only partially grounded, and 0 indicates a mismatch or an unclear direction. \emph{Minimal contrast} is captured by \textsc{PairMatch} ($Q_3 \in \{0,1,2\}$), which evaluates whether the two images remain well matched outside the core intended contrast. \emph{Shortcut suppression} is captured by \textsc{NoExplicitCue} ($Q_4 \in \{0,1\}$), which checks for explicit cues such as visible text, UI elements, flags, check marks, or cross symbols.

This rubric is used in two ways. In Section~\ref{sec:construction}, we use an automatic MLLM judge prompted with the same rubric for scalable pipeline comparison. For final benchmark inclusion, human reviewers apply the same rubric together with a strict acceptance rule, which we describe in Section~\ref{sec:human_review}.

\begin{table*}[!t]
\centering
\small
\setlength{\tabcolsep}{4.0pt}
\renewcommand{\arraystretch}{1.08}
\begin{tabular}{l|l|cccc|c}
\toprule
\raisebox{0.4\normalbaselineskip}{Backbone} 
& \raisebox{0.4\normalbaselineskip}{Configuration} 
& \shortstack{$Q_1$\\\textsc{Match-A}} 
& \shortstack{$Q_2$\\\textsc{Match-B}} 
& \shortstack{$Q_3$\\\textsc{PairMatch}} 
& \shortstack{$Q_4$\\\textsc{NoCue}} 
& \shortstack{\raisebox{0.4\normalbaselineskip}{Total Score}} \\
\midrule
GPT & \emph{Planner only} & 1.575 & 1.483 & 1.180 & 0.815 & 0.733 \\
GPT & \emph{Planner+Critic} & \textbf{1.605} & \textbf{1.508} & 1.097 & 0.749 & 0.714 \\
GPT & \emph{Planner+Editor+Critic} & 1.521 & 1.424 & \textbf{1.981} & \textbf{0.934} & \textbf{0.849} \\
\midrule
Gemini & \emph{Planner only} & \textbf{1.737} & \textbf{1.667} & 0.533 & 0.821 & 0.698 \\
Gemini & \emph{Planner+Critic} & 1.583 & 1.337 & 1.276 & 0.811 & 0.727 \\
Gemini & \emph{Planner+Editor+Critic} & 1.621 & 1.533 & \textbf{1.931} & \textbf{0.913} & \textbf{0.864} \\
\bottomrule
\end{tabular}
\caption{Automatic rubric scores from the ensemble validation judge for construction-pipeline ablations. 
Total Score is the normalized aggregate score, computed as
$(Q_1/2 + Q_2/2 + Q_3/2 + Q_4)/4$.}
\label{tab:pipeline_eval}
\end{table*}

\section{Benchmark Validation}

Before evaluating downstream MLLMs on ValueGround, we first validate the benchmark-construction process itself. The analyses in this section assess the quality of generated question-level image pairs, rather than country-conditioned benchmark accuracy on the final benchmark.

We distinguish two roles in this stage. The \emph{Critic} is an in-loop component used to revise intermediate outputs during construction, whereas the \emph{automatic validation judge} is a separate ensemble of \textsc{GPT-5.4-mini} and \textsc{Gemini 3 Flash} that scores completed candidate pairs with the four-item rubric from Section~\ref{sec:rubric}. The two judges' scores are averaged and used only for scalable comparison and screening; final benchmark filtering relies on human review.

\subsection{Construction framework comparison}
\label{sec:construction}

\paragraph{Setup.} We ablate three components of the construction pipeline: \emph{Planner}, \emph{Editor}, and \emph{Critic} for WVS Wave~7~\citep{wvs2022wave7} questions. The \emph{Planner} first expands the option pair into a shared scene plan, ensuring that the two images are generated from the same underlying scenario rather than from two independent prompts. We compare three configurations: (\textit{i}) \emph{Planner} only, which directly produces the two final image-generation prompts from the shared plan without subsequent editing or critique; (\textit{ii}) \emph{Planner}+\emph{Critic}, which adds critique-based review and regeneration; and (\textit{iii}) \emph{Planner}+\emph{Editor}+\emph{Critic}, which further introduces shared-base generation followed by targeted edits before critique. 

We instantiate each configuration under two backbone families. The GPT group uses \textsc{GPT-5.4-mini} as the MLLM agent and \textsc{GPT Image 2} as the image generation model~\citep{openai_gpt_5_4_mini_2026, openai_gpt_image_2_2026}. The Gemini group uses \textsc{Gemini 3 Flash}~\citep{google_gemini3flashpreview} as the MLLM agent and \textsc{Nano Banana 2} as the image generation model~\citep{google_nanobanana2_2026}.

\paragraph{Results.} Table~\ref{tab:pipeline_eval} shows that adding the \emph{Planner}+\emph{Editor}+\emph{Critic} yields the best total score and substantially improves pair-level control, especially on \textsc{PairMatch} ($Q_3$) and \textsc{NoExplicitCue} ($Q_4$), although \emph{Planner} only remains slightly stronger on the single-image semantic dimensions $Q_1$ and $Q_2$. Since  \textsc{ValueGround} is designed to isolate value-sensitive contrasts within tightly matched image pairs, we prioritize pair-level control over marginal gains in single-image semantic fidelity. We therefore use \emph{Planner}+\emph{Editor}+\emph{Critic} as the default construction framework, while drawing candidate pairs from both GPT and Gemini generation backbones before applying the same human acceptance rule.

\subsection{Human Review and Final Acceptance}
\label{sec:human_review}

\paragraph{Automatic judge validation.}
On a stratified 36-example subset annotated by four graduate-level annotators from China and Germany, two from each country, the automatic validation judge achieves Pearson/Spearman correlations of 0.633/0.614 with the mean normalized human total score. This is comparable to human--human agreement of 0.581/0.583. Full field-level results are provided in Appendix~\ref{app:judge_human_agreement}.

\paragraph{Final acceptance.}
For benchmark inclusion, automatically screened pairs are manually inspected and verified by the ensemble judge using the rubric in Section~\ref{sec:rubric}. A pair is retained only if
\begin{equation}
Q_1 + Q_2 \geq 3,\qquad Q_3 = 2,\qquad Q_4 = 1.
\end{equation}
This rule filters out clear semantic mismatches while enforcing strict pair-level control.

Pairs that fail the rule but can be corrected are revised by passing human review feedback to the \emph{Critic} and then rechecked under the same rubric. After automatic construction and iterative human-in-the-loop filtering on questions,  \textsc{ValueGround} retains 224 of 290 candidate questions, each with GPT- and Gemini-generated image-pair variants.
\begin{figure*}[!t]
    \centering
    \includegraphics[width=\linewidth]{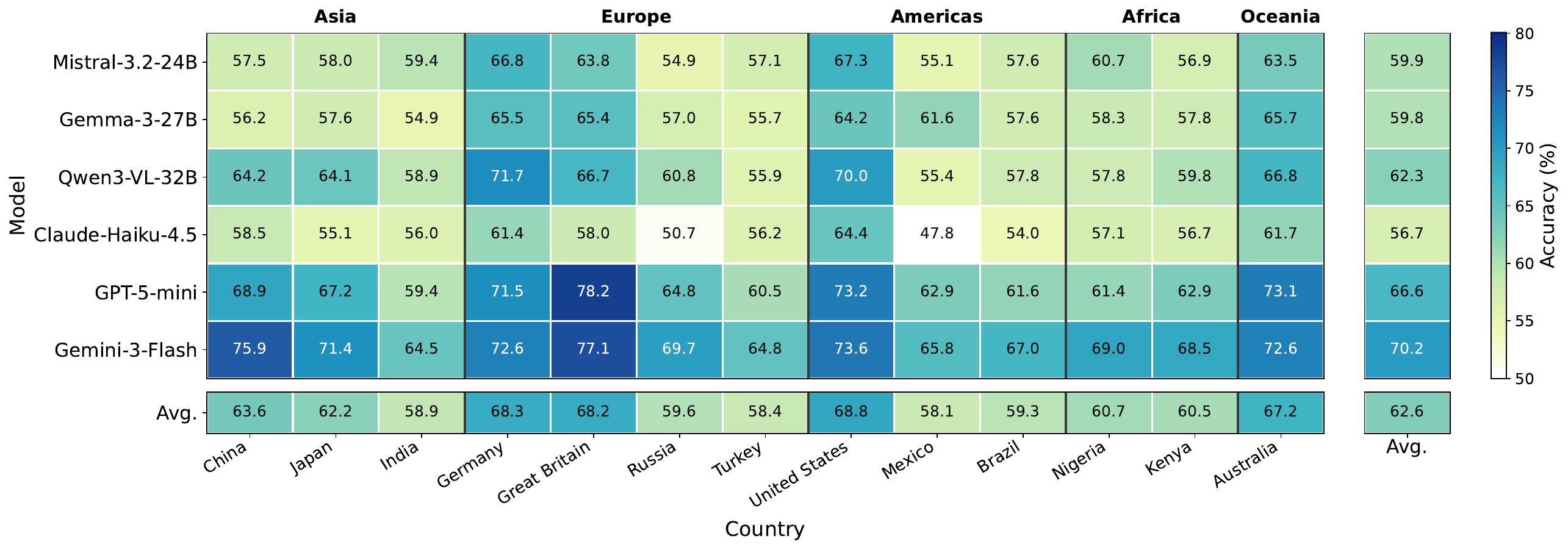}
    \caption{Country-level accuracy (\%) on the main task. Rows correspond to models and columns to countries. Darker colors indicate higher accuracy.}
    \label{fig:joint_country_heatmap}
\end{figure*}

\section{Experiments}
\label{sec:experiments}

\begin{table*}[ht]
\centering
\small
\setlength{\tabcolsep}{4.5pt}
\begin{tabular}{l|ccc|ccc}
\toprule
Model 
& Main 
& Text-only 
& Align. 
& Main$\mid$Align$\checkmark$ 
& Text$\mid$Align$\checkmark$ 
& $\Delta^{+}$ \\
\midrule
\textsc{Mistral-3.2-24B} & 59.9 & 72.8 & 87.1 & 61.1 & 72.3 & -11.1 \\
\textsc{Gemma-3-27B} & 59.8 & 68.5 & 83.7 & 60.5 & 68.6 & -8.1 \\
\textsc{Qwen3-VL-32B} & 62.3 & 72.7 & 90.2 & 63.6 & 72.7 & -9.1 \\
\textsc{Claude Haiku 4.5} & 56.7 & 70.9 & 77.5 & 59.3 & 69.9 & -10.6 \\
\textsc{GPT-5.4 mini} & 66.6 & 72.2 & 90.2 & 70.2 & 72.2 & \textbf{-1.9} \\
\textsc{Gemini 3 Flash} & \textbf{70.2} & \textbf{80.0} & \textbf{90.6} & \textbf{74.9} & \textbf{79.9} & -5.0 \\
\midrule
Avg. & 62.6 & 72.8 & 86.5 & 64.9 & 72.6 & -7.7 \\
\bottomrule
\end{tabular}
\caption{
Accuracy (\%) on the main task, text-only setting, option–image alignment setting, and alignment-filtered subsets.
\textbf{Main$\mid$Align$\checkmark$} and \textbf{Text$\mid$Align$\checkmark$} respectively report main-task and text-only accuracy on the same subset of country-question instances $(c,q)$ for which the model correctly identifies the option--image correspondence for question $q$ in the alignment task.
$\Delta^{+}$ denotes \textbf{Main$\mid$Align$\checkmark$} minus \textbf{Text$\mid$Align$\checkmark$}.
The consistently negative $\Delta^{+}$ indicates that correct option--image alignment does not eliminate the gap between text-only and main task.
}
\label{tab:decomposition_results}
\end{table*}

We instantiate \textsc{ValueGround} as a benchmark and evaluate MLLMs on culture-conditioned visual value grounding. The evaluation covers 224 English questions from WVS, 448 minimally contrastive image pairs, and country-level data from 13 countries.We first describe the evaluation setup, report the main task results, and finally analyze model behavior under auxiliary evaluation settings. Full details of prompting, output parsing, and label construction are provided in Appendix~\ref{sec:appendix:detailed-eval-protocol}.

\subsection{Evaluation Setup}
\label{sec:eval_setup}

\paragraph{Evaluation settings.}
We evaluate models in three settings. Each retained question has two validated image-pair variants, one generated by GPT and one by Gemini. All results involving images are averaged over both variants. For simplicity, we write $I_q=(I_A,I_B)$ to denote either variant.

\textbf{Main task.}
Input: $(c,q,I_q)$, where $I_q=(I_A,I_B)$.  
Output: $\hat y_{c,q}\in\{o_A,o_B\}$, indicating whether image $A$ or image $B$ better matches the value tendency associated with country $c$ for question $q$. This is the main evaluation setting.

\textbf{Binary text-only.}
Input: $(c,q,O_q)$, where $O_q=(o_A,o_B)$.
Output: $\hat y^{\text{text}}_{c,q}\in\{o_A,o_B\}$, selecting which response option better matches the same country-level value tendency. This setting isolates how much of the task can be solved from country-conditioned textual priors alone, without visual grounding.

\textbf{Option--image alignment.}
Input: $(q,O_q,I_q)$, where $O_q=(o_A,o_B)$ and $I_q=(I_A,I_B)$., without country information.
Output: $\hat m_q=(\hat m_{a},\hat m_{b})$, where $\hat m_{q,j}\in\{o_A,o_B\}$ identifies which option image $I_j$ depicts. This setting serves as a guided diagnostic for whether the model can identify the intended option--image correspondence in the constructed image pair.

\paragraph{Ground-truth labels and metrics.}
For each country--question instance $(c,q)$, we derive $y_{c,q}\in\{o_A,o_B\}$ from WVS Wave~7 by mapping the average-weighted country mean to the closer response option. Main-task and text-only results are reported as accuracy against this WVS-derived label. For option--image alignment, accuracy is computed against the known option--image correspondence and requires both images to be matched correctly.

\paragraph{Models and countries coverage.}
We evaluate the following multimodal language models: \textsc{GPT-5.4-mini}~\citep{openai_gpt_5_4_mini_2026}, \textsc{Claude Haiku 4.5}~\citep{anthropic2025haiku45}, \textsc{Gemini 3 Flash Preview}~\citep{google_gemini3flashpreview}, \textsc{Mistral-3.2-24B-Instruct-2506}~\citep{mistral_small_2506}, \textsc{Gemma-3-27B}~\citep{gemmateam2025gemma3technicalreport}, and \textsc{Qwen3-VL-32B-Instruct}~\citep{qwen_qwen3vl}. We report results on 13 countries drawn from the WVS data: Australia, Brazil, China, Germany, Great Britain, India, Japan, Kenya, Mexico, Nigeria, Russia, Turkey, and the United States.

\subsection{Results on the Main Task}
\label{sec:main_results}

\paragraph{Overall performance.}
Figure~\ref{fig:joint_country_heatmap} reports main-task accuracy on the full  \textsc{ValueGround} benchmark, averaging over all validated image-pair variants. Gemini-3-Flash achieves the best overall accuracy at 70.2\%, followed by GPT-5.4-mini at 66.6\% and Qwen3-VL-32B at 62.3\%. Mistral-3.2-24B and Gemma-3-27B obtain similar results, with 59.9\% and 59.8\%, respectively, while Claude-Haiku-4.5 performs lowest at 56.7\%. The 13.5-point gap between the strongest and weakest models indicates clear differences in culture-conditioned visual value grounding ability. At the same time, even the best-performing model remains far from perfect, suggesting that the full task remains challenging when value alternatives must be grounded in visual evidence rather than only in text.

\paragraph{Country difficulty varies substantially.}
Difficulty is not uniform across countries. Averaged across models, the United States is the easiest country at 68.8\%, followed closely by Germany at 68.3\%, Great Britain at 68.2\%, and Australia at 67.2\%. In contrast, Mexico is the hardest country at 58.1\%, followed by Turkey at 58.4\%, India at 58.9\%, Brazil at 59.3\%, and Russia at 59.6\%. The 10.7-point gap between the easiest and hardest countries shows that benchmark difficulty varies substantially across national contexts.

\paragraph{Model advantages are uneven across countries.}
Stronger models do not show uniform advantages across all countries. Gemini-3-Flash obtains the best score in 11 of 13 countries, indicating the strongest overall robustness, while GPT-5-mini achieves the highest country-level score on Great Britain, at 78.2\%, and also slightly outperforms Gemini-3-Flash on Australia, at 73.1\% versus 72.6\%. On difficult countries such as Mexico, Turkey, India, Brazil, and Russia, performance remains relatively low for most models. Overall, these results suggest that the main task requires not only general multimodal competence, but also reliable grounding of country-conditioned value tendencies in visually subtle contrasts.

\subsection{Additional Evaluation Settings}
\label{sec:additional}

We compare the main task with the two auxiliary settings introduced above to identify whether errors stem mainly from weak country-conditioned textual priors, difficulty in identifying the intended option--image correspondence, or the integration of the two.

\begin{figure*}[ht]
    \centering
    \includegraphics[width=0.95\linewidth]{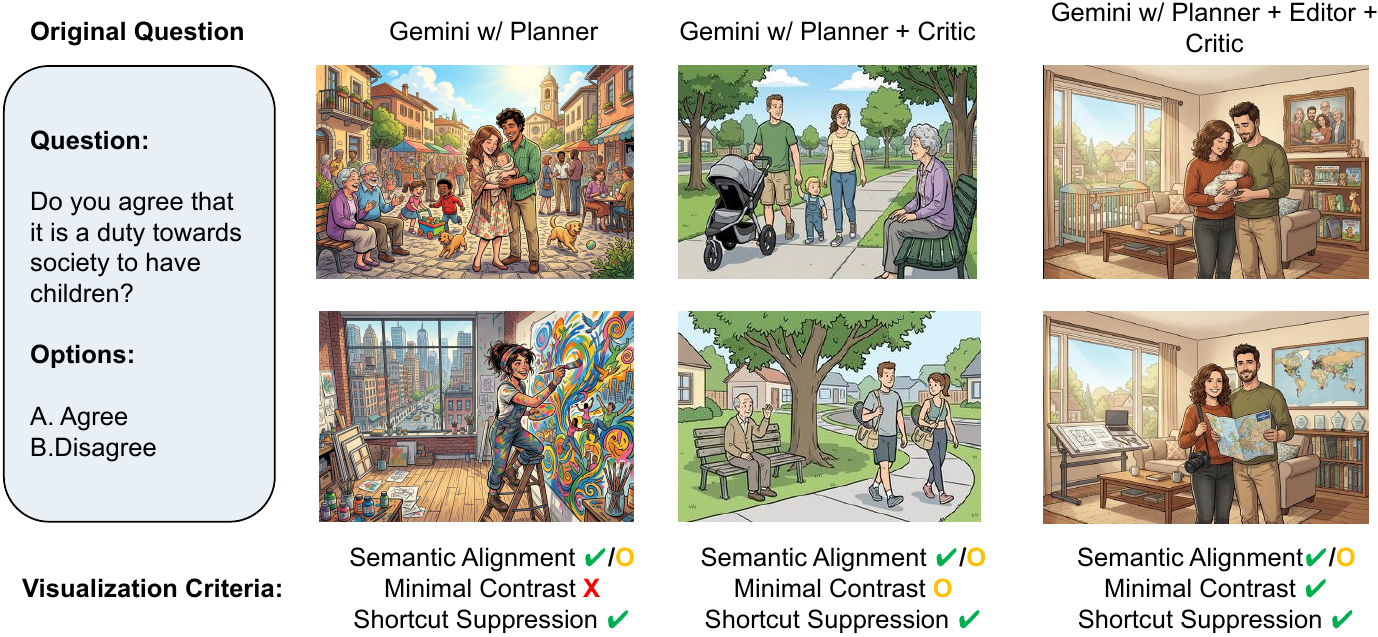}
    \caption{Illustration of the edit-based pipeline for a hard-to-visualize question. The pipeline preserves semantic alignment to the target options while producing a more controlled contrastive pair through a shared base scene and minimal edits.}
    \label{fig:qualitative_pipeline_example}
\end{figure*}

\paragraph{The main task is harder than both auxiliary settings.}
As shown in Table~\ref{tab:decomposition_results}, all models perform worse on the main task $(c,q,I_q)$ than in text-only prediction $(c,q,O_q)$ or option--image alignment $(q,O_q,I_q)$. Averaged across models, accuracy is 62.6\% on the main task, compared with 72.8\% and 86.5\% in the two auxiliary settings, indicating that the full task cannot be reduced to either textual priors or option--image matching alone.

\paragraph{Text-only tendencies transfer only partially to images.} The text-only setting suggests that models often capture useful country-conditioned verbal tendencies, but these tendencies are not reliably preserved when options are visually instantiated. The average drop from text-only to the main task is 10.2 points, with the largest decreases for Claude Haiku 4.5 and Mistral-3.2-24B. Thus, knowing the likely verbal answer is insufficient; models must also ground that answer in the paired visual alternatives.

\paragraph{Correct option--image identification is still insufficient.}
Since our image pairs are minimally contrastive, subtle visual-cue perception may contribute to the main-task drop. We therefore control for this by evaluating only alignment-correct instances, where the model identifies which image corresponds to each verbal option. Even on this subset, filtered main-task accuracy averages 64.9\%, below 72.6\% for filtered text-only accuracy, with a negative $\Delta^{+}$ for every model. Thus, visual-cue identification is not the only bottleneck; models also struggle to use the identified visual alternative for the country-conditioned WVS judgment. We provide further instance-level analysis in Appendix~\ref{sec:appendix_additional_experiment}.

\subsection{Qualitative Analysis}
\label{sec:qualitative}

Figure~\ref{fig:qualitative_pipeline_example} presents a representative construction example for the question on whether having children is a duty to society, with options \emph{Agree} and \emph{Disagree}. The notion of a \emph{duty towards society} is inherently difficult to visualize, and all three variants reflect this challenge to some extent. The key difference, however, is whether the two images form a controlled pair. Gemini w/ Planner produces plausible outputs, but the images vary simultaneously in setting, activity, and composition, making the intended contrast harder to isolate. Adding a Critic improves semantic alignment, but independent generation still leaves unnecessary variation. By contrast, the full edit-based pipeline preserves a shared base scene and expresses the contrast through minimal edits, reducing scene-level shortcuts and forcing finer-grained visual comparison. Additional qualitative examples and analysis are provided in Appendix~\ref{sec:appendix_qualitative_examples}.
\section{Conclusion}
\label{sec:conclusion}

We presented \textsc{ValueGround}, a benchmark for evaluating \emph{culture-conditioned visual value grounding} in multimodal language models. Grounded in questions from the World Values Survey, \textsc{ValueGround} combines controlled paired-image construction with automatic screening and human review, enabling a more reliable assessment of whether models ground value judgments in visual evidence under country-specific cultural context.

Experiments on six contemporary MLLMs reveal a consistent pattern. Although models often perform reasonably well when given text-only country-conditioned options, and can usually identify the intended contrast in constructed image pairs, their performance drops substantially on the full task. This gap suggests that the main difficulty lies not in textual prior knowledge or visual contrast recognition in isolation, but in integrating country context, question semantics, and visual evidence into a coherent grounded judgment. This observation is consistent with broader findings on the limitations of large language models in robust grounding and context-sensitive reasoning \cite{kostikova2025lllms}. Qualitative examples further show that model predictions can shift across modalities, indicating instability in how such value judgments are formed.

Overall, \textsc{ValueGround} provides a controlled testbed for disentangling textual priors, shallow visual cues, and genuine multimodal grounding in culture-conditioned value judgments. We hope it will support future research on multimodal grounding, cultural alignment, and more socially situated evaluation of multimodal models.

\section*{Limitations}
First, \textsc{ValueGround} reduces richer survey response distributions to binary endpoint choices. This simplification is useful for controlled construction and evaluation, but it necessarily collapses intermediate and mixed positions present in the original survey data. As discussed in Appendix~\ref{sec:appendix_midpoint_margin}, proximity to the binary midpoint is only weakly associated with cross-modal reversal, suggesting that the challenge of \textsc{ValueGround} is not merely an artifact of binarization.

Second, not all value contrasts are equally suitable for visual grounding. Some survey options are abstract, relational, or difficult to depict without introducing unintended cues. The benchmark therefore covers only the subset of culture-conditioned value contrasts that can be visualized in a controlled way. In addition, our current instantiation is based only on WVS items; whether other survey instruments can be visualized with comparable fidelity and control remains an open question.

Third, the controlled paired-image design improves internal validity at the cost of ecological realism. By enforcing shared scenes and minimal edits, the benchmark isolates the intended contrast more clearly, but real-world cultural signals are often expressed in richer, noisier, and more ambiguous settings.

Fourth, the country-conditioned labels should not be interpreted as complete accounts of any country's values. They reflect coarse survey-based tendencies at the country level rather than the full diversity of views within a population.

Finally, \textsc{ValueGround} evaluates a specific form of multimodal reasoning under controlled conditions. It should therefore be understood as a targeted diagnostic benchmark, not as a complete measure of cultural understanding or social intelligence.

\section*{Ethical Statement}
\label{sec:ethics}

\textsc{ValueGround} studies culture-conditioned value tendencies derived from survey data, but these tendencies should not be interpreted as essential, fixed, or exhaustive properties of any country or population. They reflect coarse aggregate patterns captured by a particular survey instrument and necessarily omit substantial within-country diversity as well as individual variation.

Our benchmark derives question texts and country-level value tendencies from publicly released World Values Survey (WVS) materials and uses only aggregate country-level statistics rather than participant-level records. We use WVS materials in accordance with their terms of use for non-commercial research, with proper citation, and do not redistribute the original WVS respondent-level data files. Any released benchmark artifacts are intended only for research-oriented diagnostic evaluation and should be used consistently with these access conditions.

We do not include real-person photographs or respondent-level survey data in the benchmark. In addition, visually instantiating social value contrasts carries risks of oversimplification, stereotyping, or unintended cultural essentialization. To mitigate these risks, our construction pipeline emphasizes minimal contrast, shortcut suppression, manual screening for identifiable, unsafe, or offensive content, and human review, and we avoid treating generated scenes as literal depictions of any real population. However, these safeguards can only reduce such risks, not eliminate them entirely.

We intend \textsc{ValueGround} as a diagnostic benchmark for model evaluation under controlled conditions, not as a normative statement about what any country or group ``believes'', and not as a tool for ranking cultures, profiling populations, or supporting real-world social decision-making. Annotation and validation were conducted by the authors on synthetic images and survey-derived materials; we did not conduct new human-subject data collection involving externally recruited participants.

\section*{Acknowledgments}
The NLLG group gratefully acknowledges support from the German Research Foundation (DFG) via the Heisenberg Grant EG 375/5-1.
Josif Grabocka acknowledges the funding support from the Bayerisches Landesamt für Steuern for the Bavarian AI Taxation Laboratory.
The authors gratefully acknowledge the scientific support and HPC resources provided by the Erlangen National High Performance Computing Center (NHR@FAU) of the Friedrich-Alexander-Universität Erlangen-Nürnberg (FAU). NHR@FAU hardware is partially funded by the German Research Foundation (DFG) under grant 440719683.

\bibliography{custom}

\begin{thebibliography}{40}
\providecommand{\natexlab}[1]{#1}

\bibitem[{Abdalla et~al.(2025)Abdalla, Wang, Frey, Eger, and Grabocka}]{abdalla2025zhyper}
Mohamed Hesham~Ibrahim Abdalla, Zhipin Wang, Christian Frey, Steffen Eger, and Josif Grabocka. 2025.
\newblock Zhyper: Factorized hypernetworks for conditioned llm fine-tuning.
\newblock \emph{arXiv preprint arXiv:2510.19733}.

\bibitem[{AlKhamissi et~al.(2024)AlKhamissi, ElNokrashy, Alkhamissi, and Diab}]{alkhamissi-etal-2024-investigating}
Badr AlKhamissi, Muhammad ElNokrashy, Mai Alkhamissi, and Mona Diab. 2024.
\newblock Investigating cultural alignment of large language models.
\newblock In \emph{Proceedings of the 62nd Annual Meeting of the Association for Computational Linguistics (Volume 1: Long Papers)}, pages 12404--12422.

\bibitem[{Anthropic(2025)}]{anthropic2025haiku45}
Anthropic. 2025.
\newblock \href {https://www.anthropic.com/claude-haiku-4-5-system-card} {Claude haiku 4.5 system card}.
\newblock Technical report, Anthropic.

\bibitem[{Arora et~al.(2023)Arora, Kaffee, and Augenstein}]{arora-etal-2023-probing}
Arnav Arora, Lucie-aim{\'e}e Kaffee, and Isabelle Augenstein. 2023.
\newblock \href {https://doi.org/10.18653/v1/2023.c3nlp-1.12} {Probing pre-trained language models for cross-cultural differences in values}.
\newblock In \emph{Proceedings of the First Workshop on Cross-Cultural Considerations in NLP (C3NLP)}, pages 114--130, Dubrovnik, Croatia. Association for Computational Linguistics.

\bibitem[{Bai et~al.(2025)Bai, Cai, Chen, Chen, Chen, Cheng, Deng, Ding, Gao, Ge, Ge, Guo, Huang, Huang, Huang, Hui, Jiang, Li, Li, Li, Li, Lin, Lin, Liu, Liu, Liu, Liu, Liu, Liu, Lu, Luo, Lv, Men, Meng, Ren, Ren, Song, Sun, Tang, Tu, Wan, Wang, Wang, Wang, Wang, Xie, Xu, Xu, Xu, Yang, Yang, Yang, Yang, Yu, Zhang, Zhang, Zhang, Zheng, Zhong, Zhou, Zhou, Zhou, Zhu, and Zhu}]{qwen_qwen3vl}
Shuai Bai, Yuxuan Cai, Ruizhe Chen, Keqin Chen, Xionghui Chen, Zesen Cheng, Lianghao Deng, Wei Ding, Chang Gao, Chunjiang Ge, Wenbin Ge, Zhifang Guo, Qidong Huang, Jie Huang, Fei Huang, Binyuan Hui, Shutong Jiang, Zhaohai Li, Mingsheng Li, and 45 others. 2025.
\newblock \href {https://arxiv.org/abs/2511.21631} {Qwen3-vl technical report}.
\newblock \emph{Preprint}, arXiv:2511.21631.

\bibitem[{Bandura(1977)}]{bandura1977social}
A.~Bandura. 1977.
\newblock \href {https://books.google.de/books?id=IXvuAAAAMAAJ} {\emph{Social Learning Theory}}.
\newblock Prentice-Hall series in social learning theory. Prentice Hall.

\bibitem[{Bhatia et~al.(2024)Bhatia, Ravi, Chinchure, Hwang, and Shwartz}]{bhatia-etal-2024-local}
Mehar Bhatia, Sahithya Ravi, Aditya Chinchure, EunJeong Hwang, and Vered Shwartz. 2024.
\newblock \href {https://doi.org/10.18653/v1/2024.emnlp-main.385} {From local concepts to universals: Evaluating the multicultural understanding of vision-language models}.
\newblock In \emph{Proceedings of the 2024 Conference on Empirical Methods in Natural Language Processing}, pages 6763--6782, Miami, Florida, USA. Association for Computational Linguistics.

\bibitem[{Chen et~al.(2025)Chen, Lyu, and Zhu}]{chen2025effectiveness}
Honglin Chen, Dannuo Lyu, and Liqi Zhu. 2025.
\newblock The effectiveness of social-themed picture book reading in promoting children’s prosocial behavior.
\newblock \emph{Frontiers in psychology}, 16:1569925.

\bibitem[{Durmus et~al.(2024)Durmus, Nguyen, Liao, Schiefer, Askell, Bakhtin, Chen, Hatfield-Dodds, Hernandez, Joseph, Lovitt, McCandlish, Sikder, Tamkin, Thamkul, Kaplan, Clark, and Ganguli}]{durmus-etal-2024-globalopinionqa}
Esin Durmus, Karina Nguyen, Thomas Liao, Nicholas Schiefer, Amanda Askell, Anton Bakhtin, Carol Chen, Zac Hatfield-Dodds, Danny Hernandez, Nicholas Joseph, Liane Lovitt, Sam McCandlish, Orowa Sikder, Alex Tamkin, Janel Thamkul, Jared Kaplan, Jack Clark, and Deep Ganguli. 2024.
\newblock \href {https://openreview.net/forum?id=zl16jLb91v} {Towards measuring the representation of subjective global opinions in language models}.
\newblock In \emph{Proceedings of the Conference on Language Modeling ({COLM})}.

\bibitem[{Fu et~al.(2024)Fu, Hu, Li, Feng, Wang, Lin, Roth, Smith, Ma, and Krishna}]{fu-etal-2024-blink}
Xingyu Fu, Yushi Hu, Bangzheng Li, Yu~Feng, Haoyu Wang, Xudong Lin, Dan Roth, Noah~A. Smith, Wei{-}Chiu Ma, and Ranjay Krishna. 2024.
\newblock \href {https://doi.org/10.1007/978-3-031-73337-6\_9} {{BLINK:} multimodal large language models can see but not perceive}.
\newblock In \emph{Computer Vision - {ECCV} 2024 - 18th European Conference, Milan, Italy, September 29-October 4, 2024, Proceedings, Part {XXIII}}, Lecture Notes in Computer Science, pages 148--166. Springer.

\bibitem[{Gasser et~al.(2022)Gasser, Dammert, and Murphy}]{gasser2022children}
Luciano Gasser, Yvonne Dammert, and P~Karen Murphy. 2022.
\newblock How do children socially learn from narrative fiction: Getting the lesson, simulating social worlds, or dialogic inquiry?
\newblock \emph{Educational Psychology Review}, 34(3):1445--1475.

\bibitem[{{Google}(2026{\natexlab{a}})}]{google_gemini3flashpreview}
{Google}. 2026{\natexlab{a}}.
\newblock Gemini 3 flash preview.
\newblock \url{https://ai.google.dev/gemini-api/docs/models/gemini-3-flash-preview}.
\newblock Gemini API model documentation. Accessed: 2026-03-15.

\bibitem[{{Google}(2026{\natexlab{b}})}]{google_nanobanana2_2026}
{Google}. 2026{\natexlab{b}}.
\newblock Gemini 3.1 flash image preview documentation.
\newblock \url{https://ai.google.dev/gemini-api/docs/models/gemini-3.1-flash-image-preview}.
\newblock Accessed: 2026-03-15.

\bibitem[{Haerpfer et~al.(2022)Haerpfer, Inglehart, Moreno, Welzel, Kizilova, Diez-Medrano, Lagos, Norris, Ponarin, and Puranen}]{wvs2022wave7}
C.~Haerpfer, R.~Inglehart, A.~Moreno, C.~Welzel, K.~Kizilova, J.~Diez-Medrano, M.~Lagos, P.~Norris, E.~Ponarin, and B.~Puranen. 2022.
\newblock \href {https://doi.org/10.14281/18241.24} {World values survey: Round seven -- country-pooled datafile version 6.0}.

\bibitem[{Ju et~al.(2025)Ju, Shi, Liu, Ji, Zhang, Zhang, Xu, Yang, Han, and Guo}]{ju-etal-2025-benchmarking}
Chengyi Ju, Weijie Shi, Chengzhong Liu, Jiaming Ji, Jipeng Zhang, Ruiyuan Zhang, Jiajie Xu, Yaodong Yang, Sirui Han, and Yike Guo. 2025.
\newblock \href {https://doi.org/10.18653/v1/2025.findings-acl.1028} {Benchmarking multi-national value alignment for large language models}.
\newblock In \emph{Findings of the Association for Computational Linguistics: ACL 2025}, pages 20042--20058, Vienna, Austria. Association for Computational Linguistics.

\bibitem[{Kostikova et~al.(2026)Kostikova, Wang, Bajri, Pütz, Paaßen, and Eger}]{kostikova2025lllms}
Aida Kostikova, Zhipin Wang, Deidamea Bajri, Ole Pütz, Benjamin Paaßen, and Steffen Eger. 2026.
\newblock \href {https://doi.org/10.1145/3801096} {Lllms: A data-driven survey of evolving research on limitations of large language models}.
\newblock \emph{ACM Computing Surveys}, 58(11):1–33.

\bibitem[{Li et~al.(2024)Li, Chen, Wang, Sitaram, and Xie}]{li2024culturellm}
Cheng Li, Mengzhuo Chen, Jindong Wang, Sunayana Sitaram, and Xing Xie. 2024.
\newblock \href {https://doi.org/10.52202/079017-2693} {Culturellm: Incorporating cultural differences into large language models}.
\newblock In \emph{Advances in Neural Information Processing Systems}, volume~37, pages 84799--84838. Curran Associates, Inc.

\bibitem[{Liu et~al.(2021)Liu, Bugliarello, Ponti, Reddy, Collier, and Elliott}]{liu-etal-2021-visually}
Fangyu Liu, Emanuele Bugliarello, Edoardo~Maria Ponti, Siva Reddy, Nigel Collier, and Desmond Elliott. 2021.
\newblock \href {https://doi.org/10.18653/v1/2021.emnlp-main.818} {Visually grounded reasoning across languages and cultures}.
\newblock In \emph{Proceedings of the 2021 Conference on Empirical Methods in Natural Language Processing}, pages 10467--10485, Online and Punta Cana, Dominican Republic. Association for Computational Linguistics.

\bibitem[{Liu et~al.(2024)Liu, Zhang, Xu, Shi, Jiang, Yan, Zhang, Huang, Yuan, Li, and Hu}]{liu-etal-2024-mibench}
Haowei Liu, Xi~Zhang, Haiyang Xu, Yaya Shi, Chaoya Jiang, Ming Yan, Ji~Zhang, Fei Huang, Chunfeng Yuan, Bing Li, and Weiming Hu. 2024.
\newblock \href {https://doi.org/10.18653/v1/2024.emnlp-main.1250} {{MIB}ench: Evaluating multimodal large language models over multiple images}.
\newblock In \emph{Proceedings of the 2024 Conference on Empirical Methods in Natural Language Processing}, pages 22417--22428, Miami, Florida, USA. Association for Computational Linguistics.

\bibitem[{Luo et~al.(2024)Luo, Chen, Wan, Kang, Yan, Li, Wang, Wang, Wang, Mi, Li, Ma, Sun, and Liu}]{luo-etal-2024-codis}
Fuwen Luo, Chi Chen, Zihao Wan, Zhaolu Kang, Qidong Yan, Yingjie Li, Xiaolong Wang, Siyu Wang, Ziyue Wang, Xiaoyue Mi, Peng Li, Ning Ma, Maosong Sun, and Yang Liu. 2024.
\newblock \href {https://doi.org/10.18653/v1/2024.acl-long.573} {{CODIS}: Benchmarking context-dependent visual comprehension for multimodal large language models}.
\newblock In \emph{Proceedings of the 62nd Annual Meeting of the Association for Computational Linguistics (Volume 1: Long Papers)}, pages 10639--10659, Bangkok, Thailand. Association for Computational Linguistics.

\bibitem[{{Mistral AI}(2025)}]{mistral_small_2506}
{Mistral AI}. 2025.
\newblock Mistral small 3.2.
\newblock \url{https://docs.mistral.ai/models/mistral-small-3-2-25-06}.
\newblock Mistral Docs. Open v25.06. Released: 2025-06. Accessed: 2026-03-15.

\bibitem[{Nayak et~al.(2024)Nayak, Jain, Awal, Reddy, Steenkiste, Hendricks, Stanczak, and Agrawal}]{nayak-etal-2024-benchmarking}
Shravan Nayak, Kanishk Jain, Rabiul Awal, Siva Reddy, Sjoerd~Van Steenkiste, Lisa~Anne Hendricks, Karolina Stanczak, and Aishwarya Agrawal. 2024.
\newblock \href {https://doi.org/10.18653/v1/2024.emnlp-main.329} {Benchmarking vision language models for cultural understanding}.
\newblock In \emph{Proceedings of the 2024 Conference on Empirical Methods in Natural Language Processing}, pages 5769--5790, Miami, Florida, USA. Association for Computational Linguistics.

\bibitem[{{OpenAI}(2026{\natexlab{a}})}]{openai_gpt_5_4_mini_2026}
{OpenAI}. 2026{\natexlab{a}}.
\newblock {GPT-5.4 mini Model}.
\newblock \url{https://developers.openai.com/api/docs/models/gpt-5.4-mini}.
\newblock Accessed: 2026-05-25.

\bibitem[{{OpenAI}(2026{\natexlab{b}})}]{openai_gpt_image_2_2026}
{OpenAI}. 2026{\natexlab{b}}.
\newblock {GPT Image 2 Model}.
\newblock \url{https://developers.openai.com/api/docs/models/gpt-image-2}.
\newblock Accessed: 2026-05-25.

\bibitem[{Ramaswamy et~al.(2023)Ramaswamy, Lin, Zhao, Adcock, Van Der~Maaten, Ghadiyaram, and Russakovsky}]{ramaswamy-etal-2023-geode}
Vikram~V Ramaswamy, Sing~Yu Lin, Dora Zhao, Aaron Adcock, Laurens Van Der~Maaten, Deepti Ghadiyaram, and Olga Russakovsky. 2023.
\newblock Geode: a geographically diverse evaluation dataset for object recognition.
\newblock \emph{Advances in Neural Information Processing Systems}, 36:66127--66137.

\bibitem[{Santurkar et~al.(2023)Santurkar, Durmus, Ladhak, Lee, Liang, and Hashimoto}]{santurkar2023whose}
Shibani Santurkar, Esin Durmus, Faisal Ladhak, Cinoo Lee, Percy Liang, and Tatsunori Hashimoto. 2023.
\newblock Whose opinions do language models reflect?
\newblock In \emph{International conference on machine learning}, pages 29971--30004. PMLR.

\bibitem[{Schneider et~al.(2025)Schneider, Holtermann, Biemann, and Lauscher}]{schneider-etal-2025-gimmick}
Florian Schneider, Carolin Holtermann, Chris Biemann, and Anne Lauscher. 2025.
\newblock \href {https://doi.org/10.18653/v1/2025.findings-acl.500} {{GIMMICK}: Globally inclusive multimodal multitask cultural knowledge benchmarking}.
\newblock In \emph{Findings of the Association for Computational Linguistics: ACL 2025}, pages 9605--9668, Vienna, Austria. Association for Computational Linguistics.

\bibitem[{Shi et~al.(2024)Shi, Wang, Fan, Zhang, Li, Zhang, Yin, Sheng, Qiao, and Shao}]{shi2024ch3ef}
Zhelun Shi, Zhipin Wang, Hongxing Fan, Zaibin Zhang, Lijun Li, Yongting Zhang, Zhenfei Yin, Lu~Sheng, Yu~Qiao, and Jing Shao. 2024.
\newblock \href {https://arxiv.org/abs/2403.17830} {Assessment of multimodal large language models in alignment with human values}.
\newblock \emph{Preprint}, arXiv:2403.17830.

\bibitem[{Singh et~al.(2026)Singh, Fry, Perelman, Tart, Ganesh, El-Kishky, McLaughlin, Low, Ostrow, Ananthram, Nathan, Luo, Helyar, Madry, Efremov, Spyra, Baker-Whitcomb, Beutel, Karpenko, Makelov, Neitz, Wei, Barr, Kirchmeyer, Ivanov, Christakis, Gillespie, Tam, Bennett, Wan, Huang, Sandjideh, Yang, Kumar, Saraiva, Vallone, Gheorghe, Garcia, Braunstein, Liu, Schmidt, Mereskin, Mishchenko, Applebaum, Rogerson, Rajan, Wei, Kotha, Srivastava, Agrawal, Vijayvergiya, Tyra, Nair, Nayak, Eggers, Ji, Hoover, Chen, Chen, Barak, Minaiev, Hao, Baker, Lightcap, McKinzie, Wang, Quinn, Fioca, Hsu, Yang, Yu, Zhang, Brenner, Zetino, Raymond, Lugaresi, Paz, Hudson, Whitney, Li, Chen, Cole, Voss, Ding, Shen, Huang, Colby, Hallacy, Koch, Lu, Kaplan, Kim, Minott-Henriques, Frey, Yu, Czarnecki, Reid, Wei, Decareaux, Scheau, Zhang, Forbes, Tang, Goldberg, Roberts, Palmie, Kappler, Levine, Wright, Leo, Lin, Robinson, Grabb, Chen, Lim, Salama, Bhattacharjee, Tsipras, Li, Yu, Strouse, Williams, Hunn, Bayes, Arbus, Akyurek, Le,
  Widmann, Yani, Proehl, Sert, Cheung, Schwartz, Han, Jiang, Mitchell, Sigler, Wallace, Ritter, Kavanaugh, Mays, Nikishin, Li, Such, de~Avila Belbute~Peres, Raso, Bekerman, Tsimpourlas, Chantzis, Song, Zhang, Raila, McGrath, Briggs, Yang, Parascandolo, Chabot, Kim, Zhao, Valiant, Leclerc, Salman, Wang, Sheng, Jiang, Wang, Jin, Sikchi, Schmidt, Aspegren, Chen, Qiu, Lightman, Covert, Kivlichan, Silber, Sohl, Hammoud, Clavera, Lan, Akkaya, Kostrikov, Kofman, Etinger, Singal, Hehir, Huh, Pan, Wilczynski, Pachocki, Lee, Quinn, Kiros, Kalra, Samaroo, Wang, Wolfe, Chen, Wang, Harb, Han, Wang, Zhao, Chen, Yang, Tworek, Chand, Landon, Liang, Lin, Liu, Wang, Tang, Yin, Jang, Morris, Flynn, Ferstad, Heidecke, Fishbein, Hallman, Grant, Chien, Gordon, Park, Liss, Kraaijeveld, Guay, Mo, Lawson, McGrath, Vendrow, Jiao, Lee, Steele, Wang, Mao, Chen, Hayashi, Xiao, Salahi, Wu, Sekhri, Sharma, Singhal, Li, Nguyen, Gu-Lemberg, King, Liu, Stone, Yu, Ying, Georgiev, Lim, Tirumala, Miller, Ahmad, Lv, Clare, Fauconnet, Itow, Yang,
  Romaniuk, Anise, Byron, Pathak, Maksin, Lo, Ho, Jing, Wu, Xiong, Mamitsuka, Yang, McCallum, Held, Bourgeois, Engstrom, Kuhn, Feuvrier, Zhang, Switzer, Kondraciuk, Kaiser, Joglekar, Singh, Shah, Stratta, Williams, Chen, Sun, Cayton, Li, Zhang, Aljubeh, Nichols, Haines, Schwarzer, Gupta, Shah, Guan, Huang, Dong, Wang, Glaese, Carroll, Lampe, Malek, Sharman, Zhang, Wang, Pokrass, Florian, Pavlov, Wang, Chen, Wang, Feng, Bavarian, Lin, Abdool, Rohaninejad, Soto, Staudacher, LaFontaine, Marwell, Liu, Preston, Turley, Ansman, Blades, Pancha, Mikhaylin, Felix, Handa, Rai, Keskar, Brown, Nachum, Boiko, Murk, Watkins, Gleeson, Mishkin, Lesiewicz, Baltescu, Belov, Zhokhov, Pronin, Guo, Thacker, Liu, Yuan, Liu, Dias, Puckett, Arora, Mullapudi, Gaon, Miyara, Song, Aggarwal, Marsan, Yemiru, Xiong, Kshirsagar, Nuttall, Tsiupa, Eldan, Wang, James, Ziv, Shu, Nigmatullin, Jain, Talaie, Altman, Arnesen, Toizer, Toyer, Miserendino, Agarwal, Yoo, Heon, Ethersmith, Grove, Taylor, Bubeck, Banesiu, Amdo, Zhao, Wu, Santurkar,
  Zhao, Chaudhuri, Krishnaswamy, Shuaiqi, Xia, Cheng, Anadkat, Fishman, Tobin, Fu, Jain, Mei, Egoian, Kim, Golden, Mah, Lin, Imm, Sharpe, Yadlowsky, Choudhry, Eum, Sanjeev, Khan, Stramer, Wang, Xin, Gogineni, Christianson, Sanders, Patwardhan, Degry, Shadwell, Fu, Gao, Garipov, Sriskandarajah, Sherbakov, Korbak, Kaftan, Hiratsuka, Wang, Song, Zhao, Peterson, Kharitonov, Chernova, Kosaraju, Kuo, Pong, Verma, Petrov, Jiang, Zhang, Zhou, Xie, Zhan, McCabe, DePue, Ellsworth, Bain, Thompson, Chen, Qi, Xiang, Shi, Dubois, Yu, Khakbaz, Wu, Qian, Lee, Chen, Zhang, Xiong, Tian, Cha, Bai, Yang, Yuan, Li, Zhang, Yang, Jin, Jiang, Wang, Wang, Liu, Stubenvoll, Dou, Wu, and Wang}]{singh2025openaigpt5card}
Aaditya Singh, Adam Fry, Adam Perelman, Adam Tart, Adi Ganesh, Ahmed El-Kishky, Aidan McLaughlin, Aiden Low, AJ~Ostrow, Akhila Ananthram, Akshay Nathan, Alan Luo, Alec Helyar, Aleksander Madry, Aleksandr Efremov, Aleksandra Spyra, Alex Baker-Whitcomb, Alex Beutel, Alex Karpenko, and 467 others. 2026.
\newblock \href {https://arxiv.org/abs/2601.03267} {Openai gpt-5 system card}.
\newblock \emph{Preprint}, arXiv:2601.03267.

\bibitem[{Team et~al.(2025)Team, Kamath, Ferret, Pathak, Vieillard, Merhej, Perrin, Matejovicova, Ramé, Rivière, Rouillard, Mesnard, Cideron, bastien Grill, Ramos, Yvinec, Casbon, Pot, Penchev, Liu, Visin, Kenealy, Beyer, Zhai, Tsitsulin, Busa-Fekete, Feng, Sachdeva, Coleman, Gao, Mustafa, Barr, Parisotto, Tian, Eyal, Cherry, Peter, Sinopalnikov, Bhupatiraju, Agarwal, Kazemi, Malkin, Kumar, Vilar, Brusilovsky, Luo, Steiner, Friesen, Sharma, Sharma, Gilady, Goedeckemeyer, Saade, Feng, Kolesnikov, Bendebury, Abdagic, Vadi, György, Pinto, Das, Bapna, Miech, Yang, Paterson, Shenoy, Chakrabarti, Piot, Wu, Shahriari, Petrini, Chen, Lan, Choquette-Choo, Carey, Brick, Deutsch, Eisenbud, Cattle, Cheng, Paparas, Sreepathihalli, Reid, Tran, Zelle, Noland, Huizenga, Kharitonov, Liu, Amirkhanyan, Cameron, Hashemi, Klimczak-Plucińska, Singh, Mehta, Lehri, Hazimeh, Ballantyne, Szpektor, Nardini, Pouget-Abadie, Chan, Stanton, Wieting, Lai, Orbay, Fernandez, Newlan, yeong Ji, Singh, Black, Yu, Hui, Vodrahalli, Greff, Qiu,
  Valentine, Coelho, Ritter, Hoffman, Watson, Chaturvedi, Moynihan, Ma, Babar, Noy, Byrd, Roy, Momchev, Chauhan, Sachdeva, Bunyan, Botarda, Caron, Rubenstein, Culliton, Schmid, Sessa, Xu, Stanczyk, Tafti, Shivanna, Wu, Pan, Rokni, Willoughby, Vallu, Mullins, Jerome, Smoot, Girgin, Iqbal, Reddy, Sheth, Põder, Bhatnagar, Panyam, Eiger, Zhang, Liu, Yacovone, Liechty, Kalra, Evci, Misra, Roseberry, Feinberg, Kolesnikov, Han, Kwon, Chen, Chow, Zhu, Wei, Egyed, Cotruta, Giang, Kirk, Rao, Black, Babar, Lo, Moreira, Martins, Sanseviero, Gonzalez, Gleicher, Warkentin, Mirrokni, Senter, Collins, Barral, Ghahramani, Hadsell, Matias, Sculley, Petrov, Fiedel, Shazeer, Vinyals, Dean, Hassabis, Kavukcuoglu, Farabet, Buchatskaya, Alayrac, Anil, Dmitry, Lepikhin, Borgeaud, Bachem, Joulin, Andreev, Hardin, Dadashi, and Hussenot}]{gemmateam2025gemma3technicalreport}
Gemma Team, Aishwarya Kamath, Johan Ferret, Shreya Pathak, Nino Vieillard, Ramona Merhej, Sarah Perrin, Tatiana Matejovicova, Alexandre Ramé, Morgane Rivière, Louis Rouillard, Thomas Mesnard, Geoffrey Cideron, Jean bastien Grill, Sabela Ramos, Edouard Yvinec, Michelle Casbon, Etienne Pot, Ivo Penchev, and 197 others. 2025.
\newblock \href {https://arxiv.org/abs/2503.19786} {Gemma 3 technical report}.
\newblock \emph{Preprint}, arXiv:2503.19786.

\bibitem[{Thrush et~al.(2022)Thrush, Jiang, Bartolo, Singh, Williams, Kiela, and Ross}]{thrush2022winoground}
Tristan Thrush, Ryan Jiang, Max Bartolo, Amanpreet Singh, Adina Williams, Douwe Kiela, and Candace Ross. 2022.
\newblock Winoground: Probing vision and language models for visio-linguistic compositionality.
\newblock In \emph{Proceedings of the IEEE/CVF Conference on Computer Vision and Pattern Recognition}, pages 5238--5248.

\bibitem[{Tong et~al.(2024)Tong, Liu, Zhai, Ma, LeCun, and Xie}]{tong-etal-2024-mmvp}
Shengbang Tong, Zhuang Liu, Yuexiang Zhai, Yi~Ma, Yann LeCun, and Saining Xie. 2024.
\newblock Eyes wide shut? exploring the visual shortcomings of multimodal llms.
\newblock In \emph{Proceedings of the IEEE/CVF conference on computer vision and pattern recognition}, pages 9568--9578.

\bibitem[{Vayani et~al.(2025)Vayani, Dissanayake, Watawana, Ahsan, Sasikumar, Thawakar, Ademtew, Hmaiti, Kumar, Kuckreja, Maslych, Ghallabi, Mihaylov, Qin, Shaker, Zhang, Ihsani, Esplana, Gokani, Mirkin, Singh, Srivastava, Hamerlik, Izzati, Maani, Cavada, Chim, Gupta, Manjunath, Zhumakhanova, Rabevohitra, Amirudin, Ridzuan, Kareem, More, Li, Shakya, Saad, Ghasemaghaei, Djanibekov, Azizov, Jankovic, Bhatia, Cabrera, Obando{-}Ceron, Otieno, Farestam, Rabbani, Baliah, Sanjeev, Shtanchaev, Fatima, Nguyen, Kareem, Aremu, Xavier, Bhatkal, Toyin, Chadha, Cholakkal, Anwer, Felsberg, Laaksonen, Solorio, Choudhury, Laptev, Shah, Khan, and Khan}]{vayani2025all}
Ashmal Vayani, Dinura Dissanayake, Hasindri Watawana, Noor Ahsan, Nevasini Sasikumar, Omkar Thawakar, Henok~Biadglign Ademtew, Yahya Hmaiti, Amandeep Kumar, Kartik Kuckreja, Mykola Maslych, Wafa~Al Ghallabi, Mihail~Minkov Mihaylov, Chao Qin, Abdelrahman~M. Shaker, Mike Zhang, Mahardika~Krisna Ihsani, Amiel~Gian Esplana, Monil Gokani, and 50 others. 2025.
\newblock \href {https://doi.org/10.1109/CVPR52734.2025.01822} {All languages matter: Evaluating lmms on culturally diverse 100 languages}.
\newblock In \emph{{IEEE/CVF} Conference on Computer Vision and Pattern Recognition, {CVPR} 2025, Nashville, TN, USA, June 11-15, 2025}, pages 19565--19575. Computer Vision Foundation / {IEEE}.

\bibitem[{Wang et~al.(2025)Wang, Fu, Huang, Li, Liu, Liu, Ma, Xu, Zhou, Zhang, Yan, Mo, Liu, Lu, Li, Xiao, Chang, Roth, Zhang, Poon, and Chen}]{wang-etal-2024-muirbench}
Fei Wang, Xingyu Fu, James~Y. Huang, Zekun Li, Qin Liu, Xiaogeng Liu, Mingyu~Derek Ma, Nan Xu, Wenxuan Zhou, Kai Zhang, Tianyi~Lorena Yan, Wenjie~Jacky Mo, Hsiang{-}Hui Liu, Pan Lu, Chunyuan Li, Chaowei Xiao, Kai{-}Wei Chang, Dan Roth, Sheng Zhang, and 2 others. 2025.
\newblock \href {https://openreview.net/forum?id=TrVYEZtSQH} {Muirbench: {A} comprehensive benchmark for robust multi-image understanding}.
\newblock In \emph{The Thirteenth International Conference on Learning Representations, {ICLR} 2025, Singapore, April 24-28, 2025}. OpenReview.net.

\bibitem[{Xu et~al.(2025)Xu, Leng, Yu, and Xiong}]{xu-etal-2025-self}
Shaoyang Xu, Yongqi Leng, Linhao Yu, and Deyi Xiong. 2025.
\newblock \href {https://doi.org/10.18653/v1/2025.naacl-long.350} {Self-pluralising culture alignment for large language models}.
\newblock In \emph{Proceedings of the 2025 Conference of the Nations of the Americas Chapter of the Association for Computational Linguistics: Human Language Technologies (Volume 1: Long Papers)}, pages 6859--6877, Albuquerque, New Mexico. Association for Computational Linguistics.

\bibitem[{Xu et~al.(2024)Xu, Shi, Hu, Wang, and Zhang}]{xu-etal-2024-multiskill}
Zhenran Xu, Senbao Shi, Baotian Hu, Longyue Wang, and Min Zhang. 2024.
\newblock \href {https://doi.org/10.18653/v1/2024.findings-emnlp.81} {Multiskill: Evaluating large multimodal models for fine-grained alignment skills}.
\newblock In \emph{Findings of the Association for Computational Linguistics: EMNLP 2024}, pages 1506--1523, Miami, Florida, USA. Association for Computational Linguistics.

\bibitem[{Yadav et~al.(2025)Yadav, Zhang, Hershcovich, and Shutova}]{yadav-etal-2025-beyond}
Srishti Yadav, Zhi Zhang, Daniel Hershcovich, and Ekaterina Shutova. 2025.
\newblock \href {https://doi.org/10.18653/v1/2025.findings-naacl.422} {Beyond words: Exploring cultural value sensitivity in multimodal models}.
\newblock In \emph{Findings of the Association for Computational Linguistics: NAACL 2025}, pages 7607--7623, Albuquerque, New Mexico. Association for Computational Linguistics.

\bibitem[{Zhao et~al.(2025{\natexlab{a}})Zhao, Si, Chen, Zhang, Sun, Chang, and Zhang}]{zhao-etal-2025-looking}
Haozhe Zhao, Shuzheng Si, Liang Chen, Yichi Zhang, Maosong Sun, Baobao Chang, and Minjia Zhang. 2025{\natexlab{a}}.
\newblock \href {https://doi.org/10.18653/v1/2025.emnlp-main.995} {Looking beyond text: Reducing language bias in large vision-language models via multimodal dual-attention and soft-image guidance}.
\newblock In \emph{Proceedings of the 2025 Conference on Empirical Methods in Natural Language Processing}, pages 19666--19690, Suzhou, China. Association for Computational Linguistics.

\bibitem[{Zhao et~al.(2024)Zhao, Mondal, Tandon, Dillion, Gray, and Gu}]{zhao2024worldvaluesbench}
Wenlong Zhao, Debanjan Mondal, Niket Tandon, Danica Dillion, Kurt Gray, and Yuling Gu. 2024.
\newblock Worldvaluesbench: A large-scale benchmark dataset for multi-cultural value awareness of language models.
\newblock In \emph{Proceedings of the 2024 Joint International Conference on Computational Linguistics, Language Resources and Evaluation (LREC-COLING 2024)}, pages 17696--17706.

\bibitem[{Zhao et~al.(2025{\natexlab{b}})Zhao, Ding, Zhang, Huang, Cao, Wang, Wang, Fang, Wang, Zhai, Duan, Yang, and Chen}]{zhao-etal-2025-omnialign}
Xiangyu Zhao, Shengyuan Ding, Zicheng Zhang, Haian Huang, Maosong Cao, Weiyun Wang, Jiaqi Wang, Xinyu Fang, Wenhai Wang, Guangtao Zhai, Haodong Duan, Hua Yang, and Kai Chen. 2025{\natexlab{b}}.
\newblock \href {https://doi.org/10.18653/v1/2025.acl-long.906} {{O}mni{A}lign-{V}: Towards enhanced alignment of {MLLM}s with human preference}.
\newblock In \emph{Proceedings of the 63rd Annual Meeting of the Association for Computational Linguistics (Volume 1: Long Papers)}, pages 18490--18515, Vienna, Austria. Association for Computational Linguistics.

\end{thebibliography}

\appendix
\section{Agentic Paired-Image Construction Framework Details}
\label{sec:appendix:edit-framework}

\begin{figure*}[ht]
    \centering
    \includegraphics[width=\linewidth]{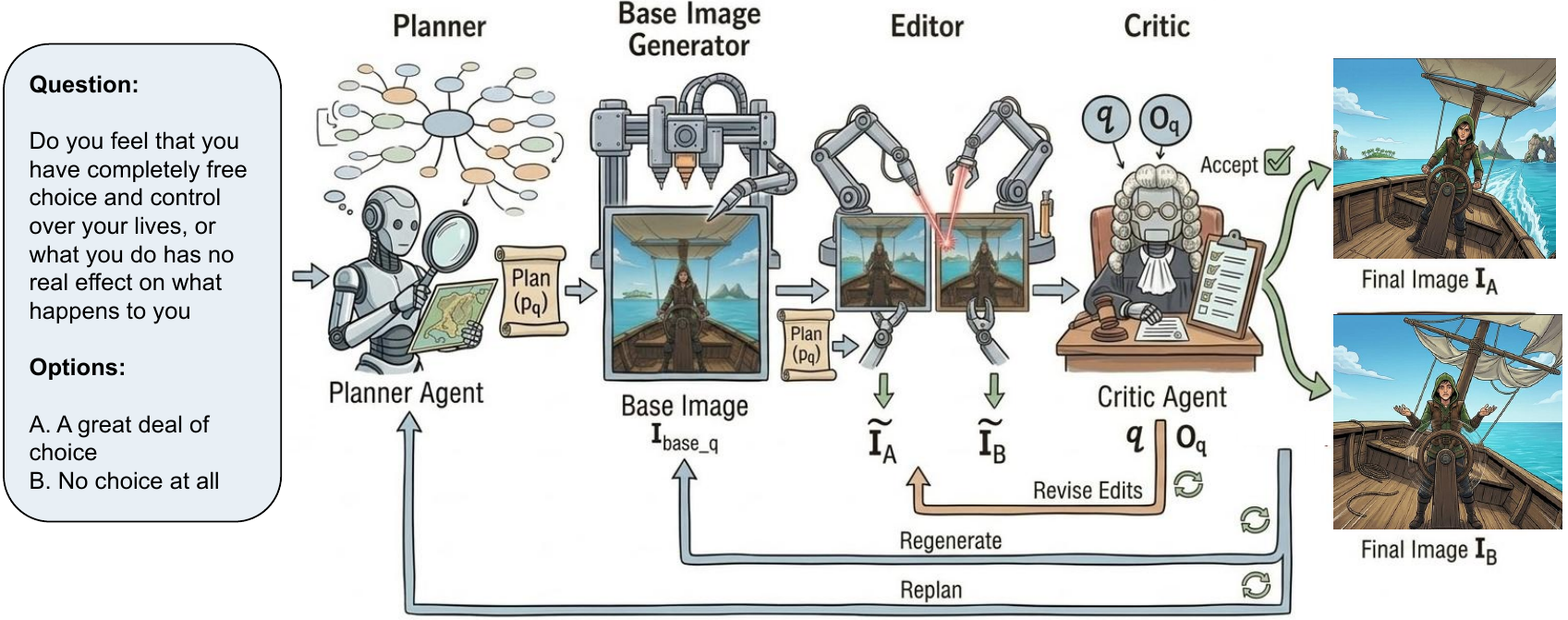}
    \caption{Overview of our agentic paired-image construction framework.}
   \label{fig:pipeline}
\end{figure*}

This appendix provides additional implementation details for the edit-based construction framework introduced in Section~\ref{sec:construction}. Given a survey question and an ordered response-option list, the framework selects two semantic endpoints, constructs a shared scene description, renders a neutral base image, edits that base image into an A/B pair, and iteratively refines the pair under critic feedback. Consistent with the main paper, the framework is organized around three coordinating agents---a \emph{Planner Agent}, an \emph{Editor Agent}, and a \emph{Critic Agent}---together with a base image generator. All language-model stages use schema-constrained JSON outputs, and all intermediate artifacts are logged for reproducibility.

\subsection{Framework Overview}

For a question $q$ with ordered response options $O_q=(o_1,\dots,o_{n_q})$, the framework first identifies a binary option pair $(o_A,o_B)$ that defines the target contrast. It then executes three agentic stages, as shown in figure~\ref{fig:pipeline}. The \emph{Planner Agent} transforms the option pair into a shared visual plan, including a common base scene, endpoint-specific signals, locked attributes that should remain invariant, editable attributes that may differ across A and B, and shortcut risks to suppress. The base image generator renders this shared scene once to obtain $I_{\text{base}}$. The \emph{Editor Agent} then produces minimal edit instructions that transform $I_{\text{base}}$ into the option images $I_A$ and $I_B$. Finally, the \emph{Critic Agent} evaluates the triplet $(I_{\text{base}}, I_A, I_B)$ and either accepts the pair or triggers another round of regeneration, edit revision, or replanning.

This design mirrors the three desiderata emphasized in the main paper: \emph{semantic alignment}, \emph{minimal contrast}, and \emph{shortcut suppression}. The planner defines these constraints at the scene level, the editor realizes them through local edits, and the critic verifies whether they are actually satisfied in the generated images.

\subsection{Structured Interfaces}

The three agents communicate through explicit structured outputs. The \emph{Planner Agent} returns three components: \texttt{endpoint\_selection}, \texttt{semantic\_statement}, and \texttt{plan}. Here \texttt{endpoint\_selection} records the chosen option pair, \texttt{semantic\_statement} rewrites each option as a standalone sentence, and \texttt{plan} specifies the visual style, shared base scene, endpoint-specific signals, locked attributes, editable attributes, and risk points. The \emph{Editor Agent} returns one edit instruction for A and one for B. The \emph{Critic Agent} returns issue fields for the three evaluation axes---semantic alignment, minimal contrast, and shortcut suppression---together with a routing decision from \{\texttt{accept}, \texttt{regenerate}, \texttt{revise\_edits}, \texttt{replan}\} and the subset of targets to revise. These structured interfaces make the roles of planning, editing, and control explicit while keeping the pipeline modular.

\subsection{Planner Agent}

\paragraph{Role.}
The \emph{Planner Agent} defines the semantic backbone of the paired visualization. Its goal is not to design two unrelated scenes, but to produce a single reusable scene that can support both option images under controlled edits. In this sense, the planner is the main component responsible for enforcing \emph{minimal contrast} at the conceptual level.

\paragraph{Inputs.}
The planner receives the question text, the full ordered option list, a banlist of shortcut cues, the current replan level, and failure feedback from earlier rounds. When raw option labels are too elliptical to support direct visual grounding, the pipeline first rewrites them into explicit option statements. For example, questionnaire-dependent labels such as \texttt{mentioned}/\texttt{not mentioned} are converted into self-contained semantic forms before planning.

\paragraph{Outputs.}
The planner outputs: (i) the selected option pair $(o_A,o_B)$, (ii) two standalone semantic statements corresponding to the two endpoints, and (iii) a structured plan containing a shared base scene, endpoint-specific signals, locked attributes, editable attributes, and risk points. Although the planner schema allows multiple styles, the current implementation fixes the final style to \texttt{comic} for greater generation stability.

\paragraph{Planning constraints.}
Planner prompts are designed to preserve the question-defining entities and relations while keeping the shared scene neutral enough to support both edits. In particular, the planner is instructed to avoid readable text, interface elements, score-like symbols, propaganda-like cues, and unrelated emotional exaggeration. It is also instructed not to rely on split-screen composition or comparison-dependent layouts; each option image should remain interpretable on its own.

\paragraph{Integrated validation.}
In the implementation, planner outputs are further screened before image generation. Rather than treating this as a separate modeling stage, we view it as part of the planner's responsibility to maintain semantic fidelity. Candidate plans are rejected if the semantic statements are incomplete, merely restate raw labels, or fail to preserve key anchor terms from the question. The planner is also prevented from producing several recurrent collapse patterns observed during development: reducing normative questions to simple object presence or absence, collapsing long-term or habitual meaning into a single momentary action, and narrowing broad everyday-importance questions into niche expert-only scenes. When such failures occur, the reasons are appended to the next planner prompt so that replanning becomes progressively more constrained.

\subsection{Base-image Generation}

The validated plan is converted deterministically into a single base prompt rather than through an additional language-model stage. This prompt combines the chosen visual style, the shared base scene, the locked attributes that should remain fixed across A and B, and the risk points that should be suppressed. The resulting image $I_{\text{base}}$ serves as the shared visual anchor for both option realizations. Because both option images are derived from the same base image, variation unrelated to the semantic contrast is reduced by construction.

\subsection{Editor Agent}

\paragraph{Role.}
The \emph{Editor Agent} translates the abstract plan into concrete image-edit instructions. Its purpose is not to redesign the scene, but to realize the endpoint-specific differences through the smallest feasible edits to the shared base image. The editor is therefore the main component that operationalizes \emph{minimal contrast} at the image level.

\paragraph{Inputs.}
The editor receives the question text, the selected endpoints, the semantic statements, the validated plan, the current base image, the previous edit instructions, critic feedback from earlier rounds, and the subset of targets that should be revised in the current round.

\paragraph{Outputs.}
It returns one edit instruction for image A and one for image B. These instructions are then sanitized before being sent to the image-edit backend in order to remove text-like or shortcut-bearing artifacts.

\paragraph{Editing constraints.}
The editor is required to preserve the shared scene, maintain the locked attributes, and modify only the editable attributes needed to express the target endpoint. It is explicitly discouraged from over-editing the scene, replacing question-defining subject cues, or introducing text, labels, logos, or interface elements. It is also instructed not to encode semantic differences through affective shortcuts unless the question itself makes emotion or attitude central. When only one side failed in the previous round, the other side is kept unchanged so as to minimize drift in already acceptable outputs.

\paragraph{Backend wrapper.}
The image-edit backend does not receive the raw editor instruction alone. Instead, the system appends a fixed wrapper that further suppresses readable text, interface-like elements, and dramatic emotional cues. This wrapper helps preserve consistent affect and discourages shortcuts that would otherwise bypass the intended semantic contrast.

\subsection{Critic Agent}

\paragraph{Role.}
The \emph{Critic Agent} controls the outer refinement loop. Given the question context, semantic targets, validated plan, edit instructions, and the image triplet $(I_{\text{base}}, I_A, I_B)$, it determines whether the current pair satisfies the construction criteria or whether another round is needed. It is the main component responsible for verifying all three desiderata jointly: \emph{semantic alignment}, \emph{minimal contrast}, and \emph{shortcut suppression}.

\paragraph{Outputs and routing decisions.}
The critic returns issue fields for each side under the three axes above and selects one routing action. \texttt{accept} is used only when both images match their option semantics, the contrast is clean and localized, and no shortcut cues remain. \texttt{regenerate} keeps the current plan and instructions but rerenders the failing side when the intended edit direction appears correct and the error is likely due to stochastic image artifacts. \texttt{revise\_edits} is used when the edit instructions are underspecified, target the wrong cue, or alter too much of the scene. \texttt{replan} is reserved for failures indicating that the shared scene or the core contrast was mis-specified at the planning stage.

\paragraph{Critic criteria.}
The critic judges semantic alignment primarily against the semantic statements rather than the raw option labels. It also checks whether the semantic difference between A and B is realized through controlled local edits rather than scene-wide divergence, and whether readable text, rating symbols, propaganda-like imagery, exaggerated emotion, or other shortcut cues remain visible. In addition, it rejects cases where key subject-defining entities or relations from the original question disappear during editing.

\paragraph{Consistency safeguard.}
To avoid logically inconsistent outcomes, the implementation automatically overrides critic decisions that conflict with the returned issue fields. In particular, a pair cannot be accepted if any issue field is non-empty. When such inconsistencies arise, semantic failures escalate the case to \texttt{replan}, whereas non-semantic failures are downgraded to \texttt{revise\_edits}.

\subsection{Retry Policy and Reproducibility}

Each planner, editor, and critic call is wrapped with retries for transient provider failures or invalid structured outputs. At the construction level, the framework permits up to two edit-revision rounds under a fixed plan and up to two full replans per question. Random seeds are derived deterministically from a base seed together with the current plan round, edit round, and target identity, making repeated runs comparable. Once edit instructions are fixed, the A and B variants can be rendered in parallel.

For reproducibility, the pipeline stores the selected endpoints, semantic statements, validated plan, base prompt, edit instructions, critic outputs, final pair metadata, and a chronological event log for each question. If the retry budgets are exhausted before acceptance, the system still writes a best-effort artifact containing the latest available images and failure notes, so that unsuccessful cases remain available for later inspection.



\subsection{Experimental Backend Configurations}

We instantiate the same edit-based construction framework with two model backends. The Gemini configuration uses \texttt{Gemini-3-flash-preview} for the planner, editor, and critic, and \textsc{Nano Banana 2} for base-image synthesis and image editing. The GPT configuration uses \texttt{GPT-5.4-mini} for the planner, editor, and critic, and \texttt{GPT-Image-2} for image generation and editing.

Both configurations use the same refinement budget. For each question, the system first attempts one planned A/B edit pair. If the critic rejects the pair, the framework allows up to two edit-level retry rounds under the same plan. These edit-level retries include both \texttt{regenerate}, which rerenders the failing target with the same edit instruction, and \texttt{revise\_edits}, which asks the editor to revise the failing edit instruction before rerendering. If the critic identifies a planning-level failure, the framework allows up to two full \texttt{replan} rounds, each of which restarts from endpoint planning, base-image generation, editor instruction generation, and critic evaluation. 

\subsection{Generation Cost of Construction Variants}
\label{sec:appendix_generation_cost}

In addition to construction quality, we report the average generation cost of the same construction variants evaluated in Table~\ref{tab:pipeline_eval}.
These costs are intended to contextualize the trade-off between generation expense and automatic validation quality; they should not be interpreted as fixed deployment costs, since absolute prices may vary with model pricing, retry frequency, and implementation details.

\begin{table}[t]
\centering
\small
\begin{tabular}{llc}
\toprule
Backbone & Configuration & Avg. cost \\
\midrule
Gemini & \emph{Planner only} & \$0.1422 \\
Gemini & \emph{Planner+Critic} & \$0.1608 \\
Gemini & \emph{Planner+Editor+Critic} & \$0.2189 \\
\midrule
GPT & \emph{Planner only} & \$0.1100 \\
GPT & \emph{Planner+Critic} & \$0.1178 \\
GPT & \emph{Planner+Editor+Critic} & \$0.2244 \\
\bottomrule
\end{tabular}
\caption{Average generation cost per candidate image pair for the construction variants evaluated in Table~\ref{tab:pipeline_eval}.}
\label{tab:generation_cost}
\end{table}

Table~\ref{tab:generation_cost} shows that the full \emph{Planner+Editor+Critic} pipeline incurs a higher average generation cost than the two ablated variants.
For Gemini, the average cost increases from \$0.1422 for \emph{Planner only} and \$0.1608 for \emph{Planner+Critic} to \$0.2189 for the full pipeline.
For GPT, the corresponding costs are \$0.1100, \$0.1178, and \$0.2244.

This additional cost is accompanied by a clear improvement in construction quality.
As shown in Table~\ref{tab:pipeline_eval}, \emph{Planner+Editor+Critic} achieves the highest Total Score for both GPT and Gemini, with especially large gains on \textsc{PairMatch}, the criterion most directly tied to minimal contrast between the two images.
Since \textsc{ValueGround} relies on minimally contrastive image pairs rather than independent visual depictions of response options, we prioritize the full edit-based pipeline despite its higher generation cost.
\section{Additional Analysis of the Automatic Validation Judge}
\label{app:judge_human_agreement}

\begin{figure*}[t]
    \centering
    \includegraphics[width=0.96\textwidth]{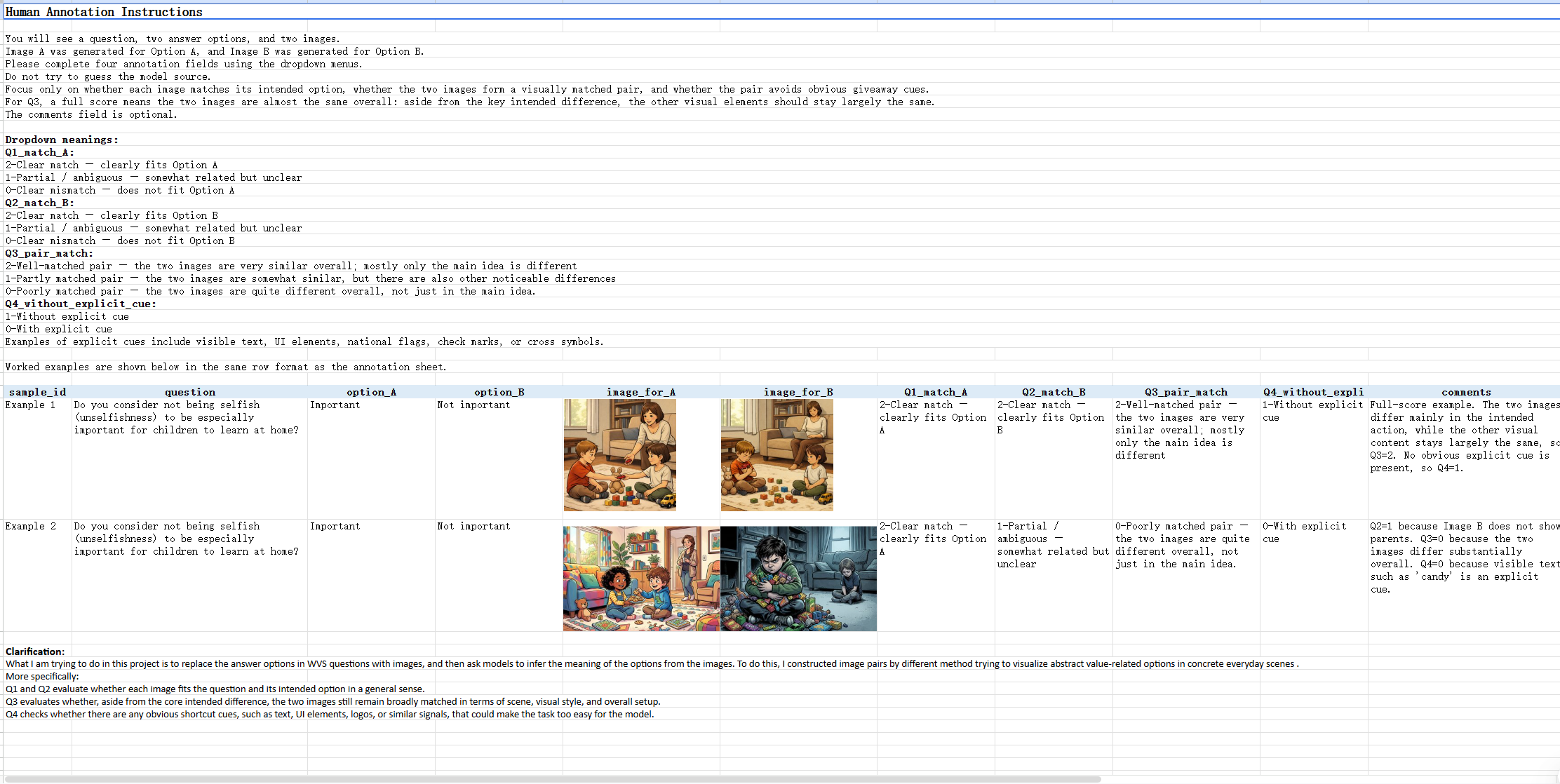}
    \caption{Human annotation instructions used for manual verification.}
    \label{fig:human_annotation_instruction}
\end{figure*}

\subsection{Rubric: Meaning of \texorpdfstring{$Q_1$--$Q_4$}{Q1--Q4} and score levels}
\label{app:rubric_details}

The image-judge rubric contains four primary questions:

\begin{itemize}
    \item \textbf{$Q_1$ (Semantic alignment with option $o_a$).}
    $Q_1$ evaluates whether Image~A is semantically closer to option $o_a$ than to the alternative option $o_b$, given the survey question.
    A score of \texttt{2} means that Image~A is both distinguishable and semantically complete: it is closer to $o_a$ and contains sufficient visual evidence for the question-option meaning.
    A score of \texttt{1} means that Image~A is distinguishable from the alternative option, but the evidence is partial, generic, or incomplete.
    A score of \texttt{0} means that Image~A is not distinguishable, matches both options equally, or better matches $o_b$.

    \item \textbf{$Q_2$ (Semantic alignment with option $o_b$).}
    $Q_2$ is defined analogously for Image~B and option $o_b$.
    A score of \texttt{2} means that Image~B clearly and completely matches $o_b$; a score of \texttt{1} means that it is closer to $o_b$ but only partially or generically expresses the intended meaning; and a score of \texttt{0} means that it is not distinguishable or better matches $o_a$.

    \item \textbf{$Q_3$ (Minimal contrast).}
    $Q_3$ evaluates the quality of the A/B pair as a controlled contrast pair rather than the semantics of either image in isolation.
    A score of \texttt{2} means that the two images are well matched overall and differ mainly in the intended semantic contrast.
    A score of \texttt{1} means that the pair is only partly matched, with noticeable extra visual differences.
    A score of \texttt{0} means that the two images differ substantially in scene, people, composition, or presentation.

    \item \textbf{$Q_4$ (Shortcut suppression).}
    $Q_4$ is a binary shortcut-control variable that asks whether the pair avoids explicit giveaway cues.
    A score of \texttt{1} means that no obvious explicit cue is present.
    A score of \texttt{0} means that the pair contains direct answer-revealing cues, such as readable text, UI elements, flags, check marks, cross symbols, country names, or other explicit symbolic labels.
\end{itemize}

\subsection{Agreement with Human Annotations}

We validate the automatic judge on a stratified 36-example subset annotated by four human reviewers using the same rubric. The automatic judge is an ensemble of \textsc{GPT-5.4-mini} and \textsc{Gemini 3 Flash}. Since the rubric fields have different scales, we compute the total score as
\begin{equation}
\mathrm{Total} = \frac{Q_1/2 + Q_2/2 + Q_3/2 + Q_4}{4}.
\end{equation}
We compare the ensemble judge against the mean human annotation. As a human agreement reference, we report leave-one-annotator-out agreement, where each annotator is compared against the mean of the other three annotators.

\begin{table}[t]
\centering
\small
\setlength{\tabcolsep}{4.5pt}
\begin{tabular}{lcccc}
\toprule
Metric & \multicolumn{2}{c}{Judge vs. Human Mean} & \multicolumn{2}{c}{Human Agreement} \\
\cmidrule(lr){2-3}\cmidrule(lr){4-5}
 & Pearson & Spearman & Pearson & Spearman \\
\midrule
$Q_1$ & 0.345 & 0.400 & 0.520 & 0.546 \\
$Q_2$ & 0.630 & 0.526 & 0.592 & 0.536 \\
$Q_3$ & 0.723 & 0.700 & 0.562 & 0.541 \\
$Q_4$ & 0.591 & 0.387 & 0.367 & 0.289 \\
Total & 0.633 & 0.614 & 0.581 & 0.583 \\
\bottomrule
\end{tabular}
\caption{Agreement between the automatic judge and human annotations on the 36-example validation subset. Human Agreement denotes leave-one-annotator-out agreement.}
\label{tab:judge_human_agreement_by_metric}
\end{table}

Table~\ref{tab:judge_human_agreement_by_metric} shows that the ensemble judge reaches a total-score Pearson correlation of 0.633 and Spearman correlation of 0.614 with the mean human annotation, comparable to leave-one-annotator-out human agreement.
\section{Detailed Evaluation Protocol}
\label{sec:appendix:detailed-eval-protocol}

This appendix supplements Section~\ref{sec:eval_setup} with details of the evaluation protocol, including WVS question filtering, prompt templates, decoding settings, input formatting, output parsing, and scoring rules. We use the same three evaluation settings and the same model set as in the main paper. 

\subsection{WVS Question Filtering}
\label{app:wvs-question-filtering}

We construct the evaluation set from the full WVS question bank. We remove
items that are not suitable for our A/B value-orientation task. Specifically,
we exclude demographic or high-cardinality variables, such as party choice,
birth year, age, country of birth, parents' country of birth, home language,
occupation, and ethnic group. We also exclude objective or behavioral-status
items, such as factual knowledge questions, organizational membership and other respondent background variables.

After this filtering, we keep subjective WVS items that ask about values,
attitudes, perceptions, evaluations, or feelings. The final evaluation set therefore
contains 224 WVS-derived questions.

\subsection{Prompt Templates}

\paragraph{Binary Text-only setting.}
The country-aware system prompt is:
\begin{quote}\footnotesize\ttfamily
You are an impartial evaluator. You will see one value-related question and two options (A/B). Choose which option better matches the typical value orientation in [COUNTRY]. Output your choice (A or B) and a brief reason.
\end{quote}

The user prompt is:
\begin{quote}\footnotesize\ttfamily
Country: [COUNTRY]\\
Question: [QUESTION]\\
Option A: [OPTION\_A]\\
Option B: [OPTION\_B]\\
Answer with: A or B, then a brief reason.
\end{quote}

\paragraph{Main task.}
The country-aware system prompt is:
\begin{quote}\footnotesize\ttfamily
You are an impartial evaluator. You will see a value-related question and two images. The two images are a contrast pair for the same question and represent opposite endpoints of that question. Choose which image (Image A or Image B) better matches the typical value orientation in [COUNTRY]. Critical rule: ignore appearance and regional cues in the images (ethnicity, clothing style, language, architecture, geography, scenery). Use only visible value signals that are relevant to the question. Output your choice (A or B, where A means Image A and B means Image B) and a brief reason.
\end{quote}

The user prompt is:
\begin{quote}\footnotesize\ttfamily
Country: [COUNTRY]\\
Question: [QUESTION]\\
The two images below are a paired comparison for opposite endpoints of this
same question.\\
Select one image as your answer.\\
Image A: [IMAGE\_A]\\
Image B: [IMAGE\_B]\\
Answer with: A or B (image choice), then a brief reason.
\end{quote}

\paragraph{Option--image alignment setting.}
This setting is country-free by construction and is used only to measure whether a model can recover the option--image correspondence for each contrastive image pair. The system prompt is:
\begin{quote}\footnotesize\ttfamily
You are an impartial image-text matching evaluator. No country prior is allowed in this task. You will receive exactly two target images. The two target images are a contrast pair for the same question and represent opposite endpoints. Match each target image to option A or option B based on option meaning and image semantics. Assign one image to A and the other image to B. Output JSON only. Include short reasons for each target image.
\end{quote}

The user prompt is:
\begin{quote}\footnotesize\ttfamily
Question: [QUESTION]\\
\\
Option A: [OPTION\_A]\\
Option B: [OPTION\_B]\\
\\
The two images below are a paired comparison for opposite endpoints of this same question.\\
\\
Target Image 1: [IMAGE\_1]\\
Target Image 2: [IMAGE\_2]\\
Return JSON only with this schema: \\
\{"image\_1":"A\_or\_B","image\_2":"A\_or\_B",\\
"reason\_1":"...","reason\_2":"..."\}.
\end{quote}

We randomly permute the two target images at evaluation time and score whether the model assigns each displayed image to the correct option. For runs without rationale generation, we use the same prompt except that the final schema is \{"image\_1":"A\_or\_B","image\_2":"A\_or\_B"\}, and the system prompt instructs the model not to include extra fields.

\subsection{Inference and Scoring Details}

We use fixed decoding settings intended to reduce generation stochasticity. The evaluated models are \textsc{GPT-5.4 mini}, \textsc{Claude Haiku 4.5}, \textsc{Gemini 3 Flash Preview}, \textsc{Mistral Small 3.2 24B Instruct 2506}, \textsc{Gemma 3 27B}, and \textsc{Qwen3-VL-32B-Instruct}. The local open-source models (\textsc{Mistral Small 3.2 24B Instruct 2506}, \textsc{Gemma 3 27B}, and \textsc{Qwen3-VL-32B-Instruct}) are run on a single H100 GPU with sampling disabled and a maximum of 128 new tokens. The API-based models (\textsc{GPT-5.4 mini}, \textsc{Claude Haiku 4.5}, and \textsc{Gemini 3 Flash Preview}) use a maximum output budget of 2048 tokens; temperature is set to 0 for \textsc{Claude Haiku 4.5} and \textsc{Gemini 3 Flash Preview}, while the \textsc{GPT-5.4 mini} path constrains output length but does not explicitly set temperature. In the reported runs, image pairs are provided as two separate image inputs.

In the reported \textbf{main task} and \textbf{option--image alignment setting}, the two images are provided as two
separate image inputs rather than as a single stitched panel. Models are asked to provide a brief rationale, but only the parsed discrete decision is used for scoring.

For the \textbf{binary text-only setting} and the \textbf{main task}, outputs are parsed as either \texttt{A} or \texttt{B}. In the \textbf{main task}, this image-level choice is then mapped back to the corresponding option label in $\{o_A,o_B\}$ for scoring.

For the \textbf{option--image alignment setting}, outputs are first parsed as a one-to-one JSON assignment from \texttt{image\_1} and \texttt{image\_2} to \texttt{A}/\texttt{B}. This assignment is then converted into the binary label space $\{\textsc{Aligned}, \textsc{Swapped}\}$. Specifically, if the predicted assignment matches the displayed order $(I_A,I_B)\mapsto(o_A,o_B)$, the example is scored as \textsc{Aligned}; otherwise it is scored as \textsc{Swapped}. Outputs that cannot be parsed are marked as unscorable.

To control for positional bias, option order in the \textbf{binary text-only setting} and image order in the \textbf{main task} and \textbf{option--image alignment setting} are randomized per question. Predictions are then mapped back to the canonical option order before final scoring.

\subsection{Binary Label Construction}
\label{sec:appendix:binary-label-construction}

\paragraph{Country-conditioned labels for the Binary Text-only setting and the Main task.}
For the \textbf{Binary Text-only setting} and the \textbf{Main task}, the target is the same country-conditioned binary label $y_{c,q}\in\{o_A,o_B\}$ derived from the binary option pair $O_q=(o_A,o_B)$ defined in Section~\ref{sec:setup}. Let $R_{c,q}$ denote the set of valid ordinal response codes for country $c$ on question $q$, and let $n_r$ be the response count for code $r\in{R}_{c,q}$. We treat only positive response codes as valid ordinal answers and compute the empirical mean response
\begin{equation}
\bar u_{c,q}
=
\frac{\sum_{r\in R_{c,q}} r\,n_r}
     {\sum_{r\in R_{c,q}} n_r}.
\end{equation}

\paragraph{Resolving option codes.}
Let $r_A$ and $r_B$ denote the numeric response codes associated with $o_A$ and $o_B$, respectively. When explicit option codes are available in the question metadata, we use them directly. Otherwise, we recover them from numeric prefixes in the option labels when such prefixes are present. If neither source is available, we fall back to the minimum and maximum valid response codes observed for that question.

\paragraph{Nearest-option assignment.}
The ground-truth country-conditioned binary label is determined by which option code is closer to the empirical mean:
\begin{equation}
y_{c,q}
=
\begin{cases}
o_A & \text{if } |\bar u_{c,q}-r_A| < |\bar u_{c,q}-r_B|,\\
o_B & \text{if } |\bar u_{c,q}-r_B| < |\bar u_{c,q}-r_A|,\\
\texttt{TIE} & \text{otherwise.}
\end{cases}
\end{equation}
The corresponding midpoint margin $|u_{c,q}-0.5|$ provides a natural measure of how close a country-question pair is to the binary decision boundary. We later analyze its relation to cross-modal reversal in Appendix~\ref{sec:appendix_midpoint_margin}. Examples with missing response counts, unresolved option codes, or exact midpoint ties are treated as unscorable and excluded from the accuracy denominator.

\section{Additional Experiment}
\label{sec:appendix_additional_experiment}

\subsection{Instance-level Cross-modal Reversal}
\label{sec:appendix_reversal}

\paragraph{Evaluation Setup.}
We further quantify how often a model changes its country-conditioned decision when moving from the \textbf{text-only} setting to the \textbf{main task}.
In the final evaluation, we compute this quantity over source-specific country-question instances. Let $s$ denote the image source, and consider each instance $(c,q,s)$ that is scorable in both settings. We denote the canonical prediction in the text-only setting as
$\hat{y}^{\text{text}}_{c,q}\in\{o_A,o_B\}$, and the canonical prediction in the main task as
$\hat{y}^{\text{main}}_{c,q,s}\in\{o_A,o_B\}$, after undoing any option-order or image-order randomization. We define a \emph{cross-modal reversal} as
\begin{equation}
\hat{y}^{\text{text}}_{c,q} \neq \hat{y}^{\text{main}}_{c,q,s}.
\end{equation}

Let $y_{c,q}\in\{o_A,o_B\}$ denote the gold country-level label.
We distinguish two directional cases:
\begin{equation}
\text{harmful: } \hat{y}^{\text{text}}_{c,q}=y_{c,q}
\ \text{and}\
\hat{y}^{\text{main}}_{c,q,s}\neq y_{c,q},
\end{equation}
\begin{equation}
\text{beneficial: } \hat{y}^{\text{text}}_{c,q}\neq y_{c,q}
\ \text{and}\
\hat{y}^{\text{main}}_{c,q,s}=y_{c,q}.
\end{equation}
Because the decision space is binary, every reversal belongs to exactly one of these two categories.

\begin{table}[t]
\centering
\small
\begin{tabular}{lccc}
\toprule
Model & Reversal & Harmful & Beneficial \\
\midrule
\textsc{Qwen3-VL-32B} & 29.7 & 21.2 & 8.6 \\
\textsc{Mistral-3.2-24B} & 34.0 & 24.3 & 9.7 \\
\textsc{Gemma-3-27B} & 35.0 & 22.5 & 12.5 \\
\textsc{Claude Haiku 4.5} & 38.7 & 26.8 & 11.8 \\
\textsc{GPT-5.4 mini} & 22.0 & 14.8 & 7.2 \\
\textsc{Gemini 3 Flash} & 22.5 & 17.4 & 5.1 \\
\bottomrule
\end{tabular}
\caption{Cross-modal reversal rates (\%) in the final evaluation. Percentages are computed over $5{,}730$ source-specific country-question instances per model. Harmful and beneficial reversals sum to the total reversal rate up to rounding.}
\label{tab:cross-modal-reversal-main}
\end{table}

\paragraph{Main result.}
Table~\ref{tab:cross-modal-reversal-main} summarizes cross-modal reversal rates in the final evaluation. For each model, we restrict the analysis to source-specific country-question instances for which the gold label is uniquely defined and the model outputs in both settings can be mapped to canonical A/B choices, yielding $N=5{,}730$ instances per model.

Cross-modal reversals are common across all six models.
The lowest reversal rates are observed for \textsc{GPT-5.4 mini} (22.0\%) and \textsc{Gemini 3 Flash} (22.5\%), whereas \textsc{Claude Haiku 4.5} shows the highest reversal rate at 38.7\%. In all six models, the harmful reversal rate exceeds the beneficial reversal rate. Thus, moving from text-only evaluation to the main task more often turns a correct text-only decision into an incorrect visual decision than it corrects a text-only mistake. Correspondingly, the difference between the harmful and beneficial rates equals the model's accuracy drop from text-only to the main task on the same set of source-specific evaluation instances, up to rounding.

\begin{table}[t]
\centering
\small
\begin{tabular}{lccc}
\toprule
Country & Reversal & Harmful & Beneficial \\
\midrule
Russia & 33.5 & 22.3 & 11.2 \\
Japan & 32.9 & 22.2 & 10.7 \\
Mexico & 32.0 & 22.7 & 9.3 \\
China & 31.9 & 21.3 & 10.6 \\
Turkey & 31.6 & 20.9 & 10.7 \\
Kenya & 30.5 & 20.8 & 9.7 \\
India & 30.4 & 19.7 & 10.7 \\
Brazil & 30.1 & 21.4 & 8.7 \\
Nigeria & 29.1 & 19.7 & 9.4 \\
Great Britain & 28.9 & 22.0 & 6.9 \\
United States & 28.4 & 21.8 & 6.6 \\
Australia & 28.0 & 21.1 & 6.9 \\
Germany & 27.0 & 19.5 & 7.5 \\
\bottomrule
\end{tabular}
\caption{Country-level cross-modal reversal rates (\%) in the final evaluation. Percentages are computed by aggregating all available model-source estimates for each country. Harmful and beneficial reversals sum to the total reversal rate up to rounding.}
\label{tab:cross-modal-reversal-country}
\end{table}

\paragraph{Country-level pattern.}
Table~\ref{tab:cross-modal-reversal-country} reports country-level reversal rates aggregated over all available model-source estimates. Reversals occur consistently across countries, ranging from 27.0\% for Germany to 33.5\% for Russia. The harmful rate exceeds the beneficial rate for every country, showing that the cross-modal shift systematically reduces agreement with the gold country-level labels rather than merely changing predictions in both directions equally. This suggests that the observed text-to-image performance gap is not driven by a small subset of countries, but reflects a broader instability in transferring country-conditioned value judgments from textual options to visualized alternatives.

\subsection{Relation Between Cross-modal Reversal and WVS Midpoint Margin}
\label{sec:appendix_midpoint_margin}

\paragraph{Setup.}
We next examine whether cross-modal reversal is related to the strength of the empirical WVS preference underlying each binary country-level label.
For each country-question pair $(c,q)$, let $u_{c,q}\in[0,1]$ denote the normalized weighted WVS mean used for binary label assignment, and define the \emph{midpoint margin}
\begin{equation}
d_{c,q}=|u_{c,q}-0.5|.
\end{equation}
Smaller $d_{c,q}$ indicates that the country-question pair lies closer to the binary decision boundary, and therefore reflects a weaker empirical preference between $o_A$ and $o_B$.

In the final evaluation, reversal is computed over source-specific observations. For each model $m$, country $c$, question $q$, and image source $s$, we define
\begin{equation}
r_{m,c,q,s}
=
\mathbf{1}\!\left[
\hat y^{\text{text}}_{m,c,q}\neq \hat y^{\text{main}}_{m,c,q,s}
\right],
\end{equation}
where $\hat y^{\text{text}}_{m,c,q}$ is the canonical text-only prediction and $\hat y^{\text{main}}_{m,c,q,s}$ is the corresponding main-task prediction for source $s$.

\begin{table}[t]
\centering
\small
\begin{tabular}{lcc}
\toprule
Bin & Midpoint-margin range & Reversal rate \\
\midrule
Bin 1 & $[0.000,\,0.061)$ & 33.3 \\
Bin 2 & $[0.061,\,0.126)$ & 30.6 \\
Bin 3 & $[0.126,\,0.205)$ & 29.6 \\
Bin 4 & $[0.205,\,0.304)$ & 28.0 \\
Bin 5 & $[0.304,\,0.497]$ & 30.1 \\
\bottomrule
\end{tabular}
\caption{Cross-modal reversal rate (\%) as a function of midpoint margin
$d_{c,q}=|u_{c,q}-0.5|$.
Bins are equal-frequency bins over the 2{,}865 country-question pairs; each bin contains
573 country-question pairs and 6{,}876 scorable model-country-question-source observations from the
final evaluation.
Smaller midpoint margin indicates that the country-question pair lies closer to the binary WVS
decision boundary.}
\label{tab:reversal-vs-midpoint-margin}
\end{table}

\paragraph{Result.}
Table~\ref{tab:reversal-vs-midpoint-margin} reports reversal rates after grouping country-question pairs by midpoint margin.
The lowest-margin bin has the highest reversal rate, at 33.3\%, suggesting that instances closer to the WVS binary boundary are more prone to cross-modal prediction changes.
Reversal rates decrease from Bin~1 to Bin~4, but increase again in the highest-margin bin, reaching 30.1\%.
Thus, midpoint margin is related to cross-modal reversal, but the relationship is not strictly monotonic.

Overall, these results suggest that boundary ambiguity explains part, but not all, of the discrepancy between the text-only setting and the main task.
Country-question pairs with weaker empirical WVS preferences are especially likely to exhibit reversals, but reversals remain common even for larger-margin pairs.
This indicates that the performance gap is not merely a consequence of labels being close to the binary decision boundary.
Rather, replacing verbal response options with paired visual realizations introduces additional difficulty for transferring country-conditioned value judgments across modalities.

\subsection{Country-level Transfer from Text-only to the Main Task}
\label{sec:appendix_text_full_corr}

\paragraph{Evaluation Setup.}
To test whether strong performance in the text-only setting transfers to the main task, we pair, for each model, its text-only and main-task accuracies over the same 13 evaluation countries in the final evaluation.
For the main task, country-level accuracy is computed by aggregating over all scorable source-specific observations for each model-country pair.
This yields 13 paired country-level points per model and 78 pooled model-country points overall.
We do not include the country-context ablation in this analysis because it contains only main-task evaluations and therefore provides no matched text-only counterpart.

\begin{table*}[t]
\centering
\small
\begin{tabular}{lcccc}
\toprule
Model & Pearson $r$ & Spearman $\rho$ & Mean $\Delta$ (Main$-$Text) & Main $<$ Text \\
\midrule
\textsc{Qwen3-VL-32B} & 0.908 & 0.801 & $-12.6$ & 13/13 \\
\textsc{Mistral-3.2-24B} & 0.839 & 0.736 & $-14.7$ & 13/13 \\
\textsc{Gemma-3-27B} & 0.833 & 0.841 & $-10.0$ & 13/13 \\
\textsc{Claude Haiku 4.5} & 0.548 & 0.632 & $-15.0$ & 13/13 \\
\textsc{GPT-5.4 mini} & 0.937 & 0.901 & $-7.6$ & 13/13 \\
\textsc{Gemini 3 Flash} & 0.899 & 0.834 & $-12.3$ & 13/13 \\
\midrule
Pooled (78 points) & 0.792 & 0.769 & $-12.0$ & 78/78 \\
\bottomrule
\end{tabular}
\caption{Country-level correlation between text-only and main-task accuracy. Each model contributes 13 paired country-level points from the final evaluation. Mean $\Delta$ (Main$-$Text) denotes main-task accuracy minus text-only accuracy in percentage points. The last column reports the number of countries for which main-task accuracy is lower than text-only accuracy.}
\label{tab:text-vs-full-correlation}
\end{table*}

\begin{figure}[t]
    \centering
    \includegraphics[width=0.9\linewidth]{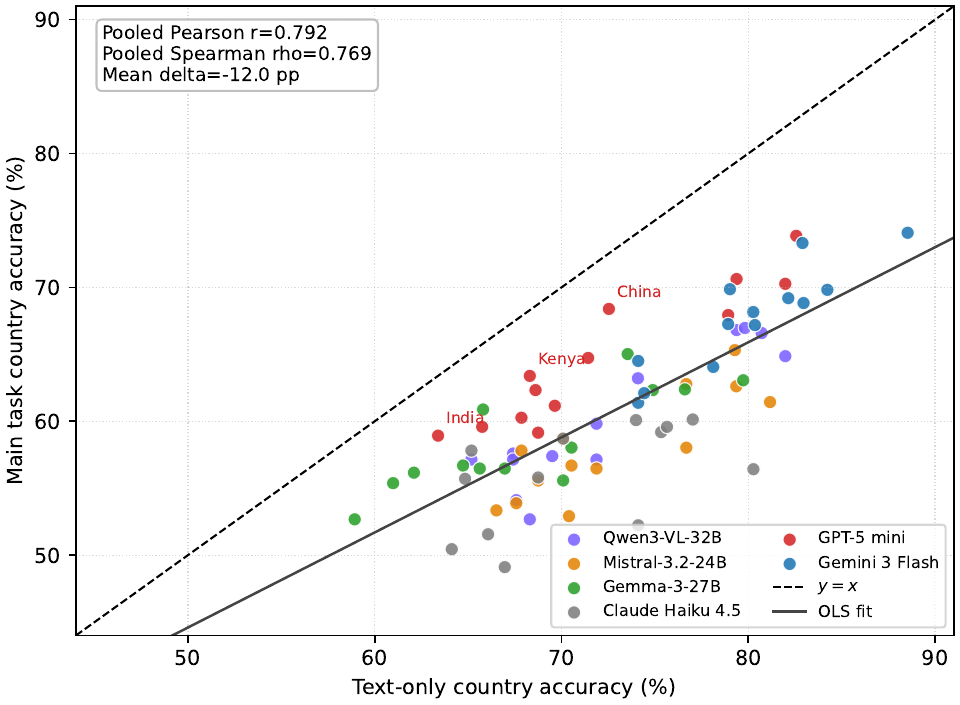}
    \caption{Country-level text-only versus main-task accuracy across all 78 model-country pairs.
    Colors indicate models.
    The dashed line denotes $y=x$, and the solid line is the least-squares fit.
    Although the two settings are strongly positively correlated, all points lie below the diagonal,
    indicating that main-task accuracy is consistently lower than text-only accuracy.}
    \label{fig:text_vs_full_scatter}
\end{figure}

\paragraph{Results.}
Figure~\ref{fig:text_vs_full_scatter} and Table~\ref{tab:text-vs-full-correlation} show a strong positive relationship between text-only and main-task performance.
For all six models, the country-level Pearson correlation between the two settings is positive, ranging from 0.548 to 0.937, with a median of 0.869.
The corresponding Spearman correlations range from 0.632 to 0.901, with a median of 0.818.
Pooling all 78 model-country points yields Pearson $r=0.792$ and Spearman $\rho=0.769$.

At the same time, the positive correlation does not imply that text-only performance transfers without loss.
Main-task accuracy is lower than text-only accuracy for every model-country pair, i.e., 78 out of 78 pooled points.
The mean main-task minus text-only gap is $-12.0$ percentage points overall, ranging from $-7.6$ points for \textsc{GPT-5.4 mini} to $-15.0$ points for \textsc{Claude Haiku 4.5}.
Thus, countries that are easier in the text-only setting also tend to be easier in the main task, but the visualized setting consistently lowers performance. 

Overall, these results suggest that country-conditioned textual knowledge provides a useful prior, but it does not reliably transfer to paired-image response options.
This supports our interpretation that \textsc{ValueGround} introduces a visual-grounding bottleneck beyond country-conditioned text-only prediction.

\section{Additional Qualitative Examples}
\label{sec:appendix_qualitative_examples}

\subsection{Construction Examples}
\label{sec:appendix_construction_examples}

\begin{figure*}[t]
    \centering
    \includegraphics[width=\textwidth]{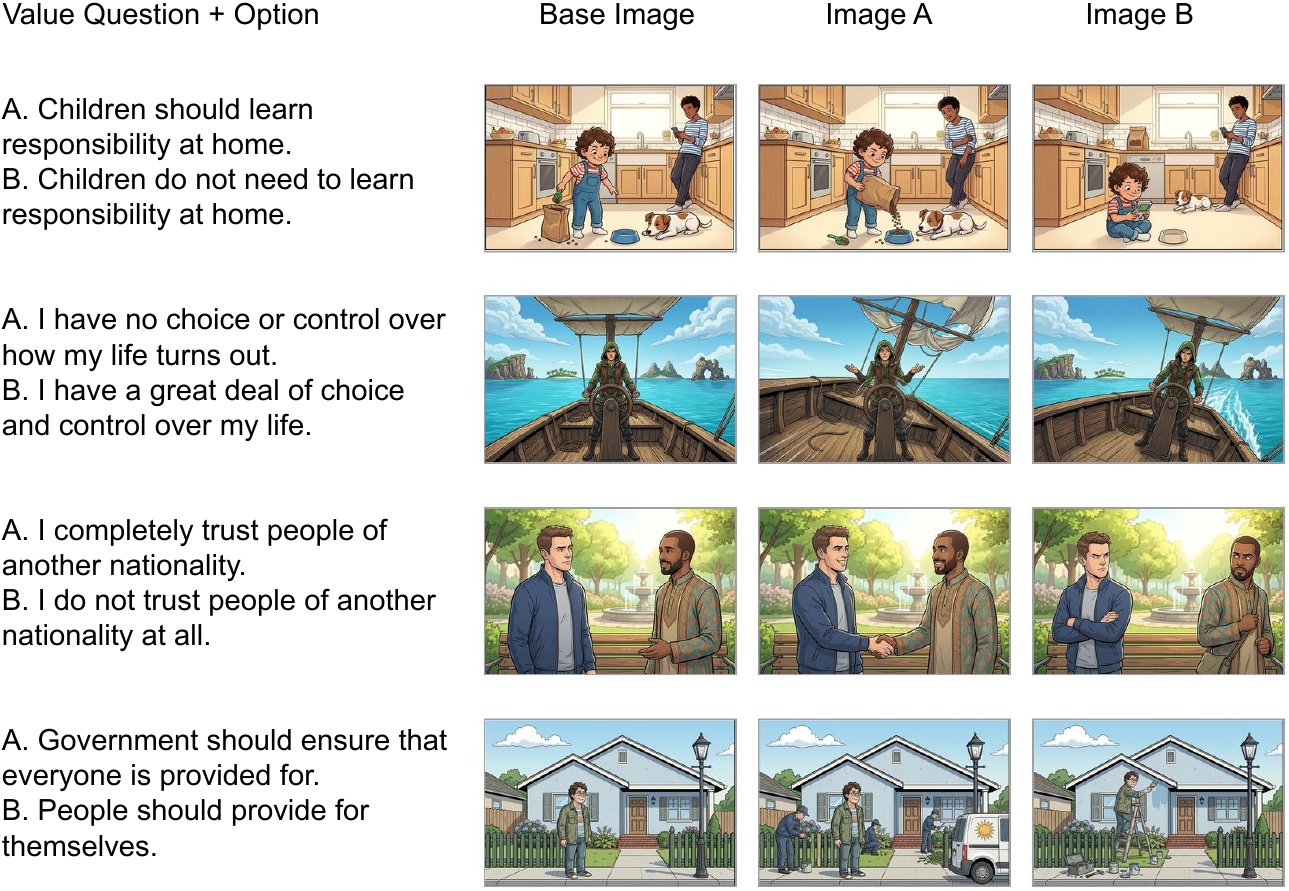}
    \caption{Additional qualitative examples from the edit-based paired-image construction pipeline.
    Each row shows a value question, the shared base image, and the final edited pair $(I_A, I_B)$.
    The examples illustrate how option-level value contrasts are realized through controlled edits on a common scene while preserving overall layout, participants, framing, and style.}
    \label{fig:appendix_construction_examples}
\end{figure*}

Figure~\ref{fig:appendix_construction_examples} shows additional qualitative examples from the edit-based paired-image construction pipeline.
Each row contains a value question with its binary option pair, the shared base image, and the final edited pair $(I_A, I_B)$.
The examples illustrate how abstract value contrasts are instantiated through localized edits on a common scene.

Specifically, family responsibility is grounded in whether the child performs household chores, such as feeding the dog.
Perceived choice and control over life is expressed through whether the person is stranded on a tilted boat turning an ineffective wheel, or instead actively steering and moving forward through the waves.
Trust toward people of another nationality is conveyed through interpersonal interaction and body language.
The contrast between public provision and self-reliance is represented through the presence or absence of external support.

Across all cases, the base scene, participants, framing, and visual style are largely preserved, while the contrast is concentrated on the option-defining attributes.
This design reduces irrelevant variation and helps ensure that performance depends on grounding the intended value difference rather than exploiting superficial scene cues.

\subsection{Iterative Refinement Example}
\label{sec:appendix_iterative_refinement_example}

\begin{figure*}[t]
    \centering
    \includegraphics[width=\textwidth]{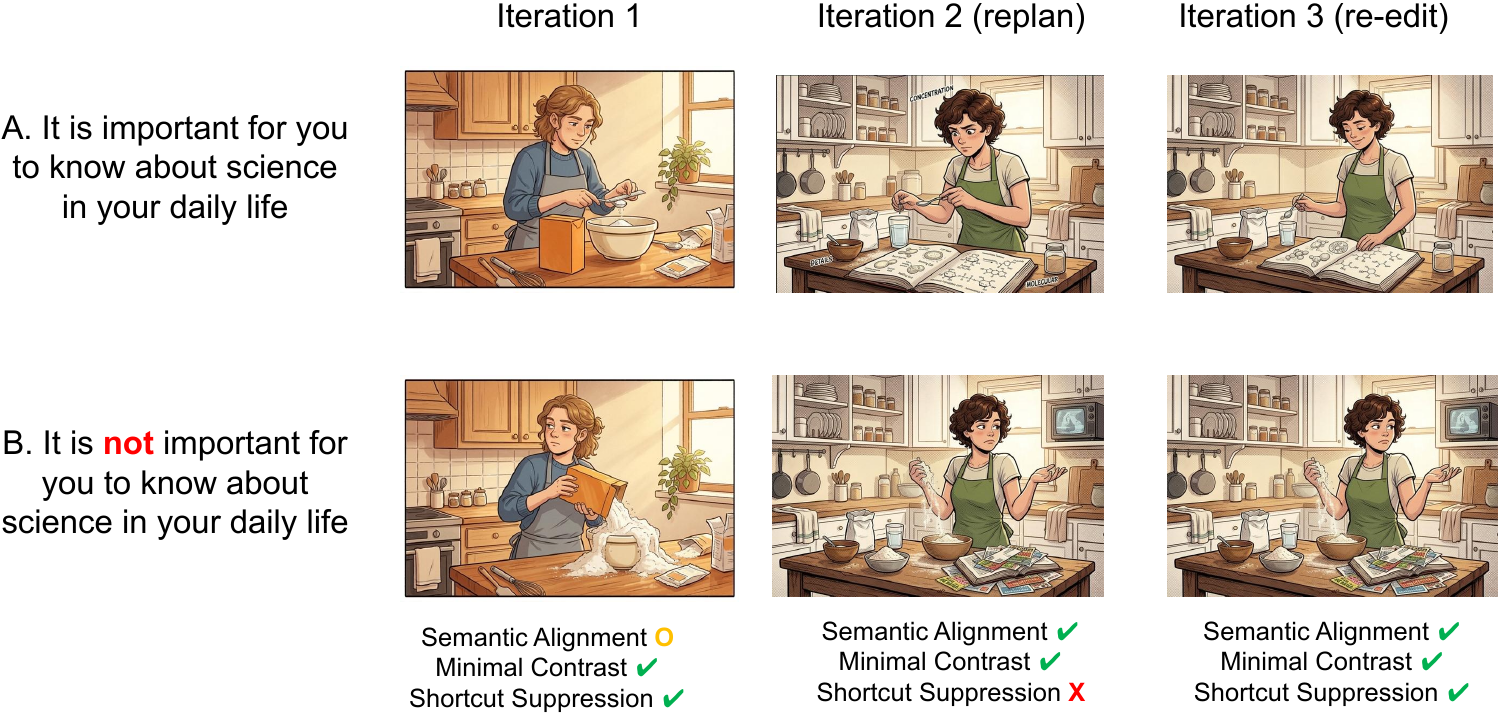}
    \caption{A representative iterative refinement example from the quality-control loop.
    Iteration 1 fails semantic alignment.
    Iteration 2, obtained after replanning, resolves the semantic issue but still violates shortcut suppression because readable textual details appear in the images.
    Iteration 3, obtained after re-editing, removes the textual details and satisfies semantic alignment, minimal contrast, and shortcut suppression, and is therefore accepted.}
    \label{fig:appendix_iterative_refinement_example}
\end{figure*}

Figure~\ref{fig:appendix_iterative_refinement_example} shows a representative example from the iterative quality-control loop.
For the question on whether it is important to know about science in daily life, the first attempt does not yet provide a sufficiently clean semantic realization of the intended contrast and therefore fails semantic alignment.
The second attempt is obtained after \emph{replan}, which revises the scene specification and improves the semantic contrast, but the resulting images still contain readable textual details and therefore fail shortcut suppression.
The third attempt is produced through \emph{re-edit}, which preserves the revised plan while removing the textual details; the final pair satisfies semantic alignment, minimal contrast, and shortcut suppression.

This example shows that the pipeline is not a one-shot generation procedure.
Instead, critic feedback can trigger either plan-level revision or edit-level correction until all acceptance criteria are satisfied.
The resulting refinement process improves both semantic faithfulness and pair quality, while keeping the contrast localized to the intended value dimension.

\begin{figure*}[t]
    \centering
    \includegraphics[width=\textwidth]{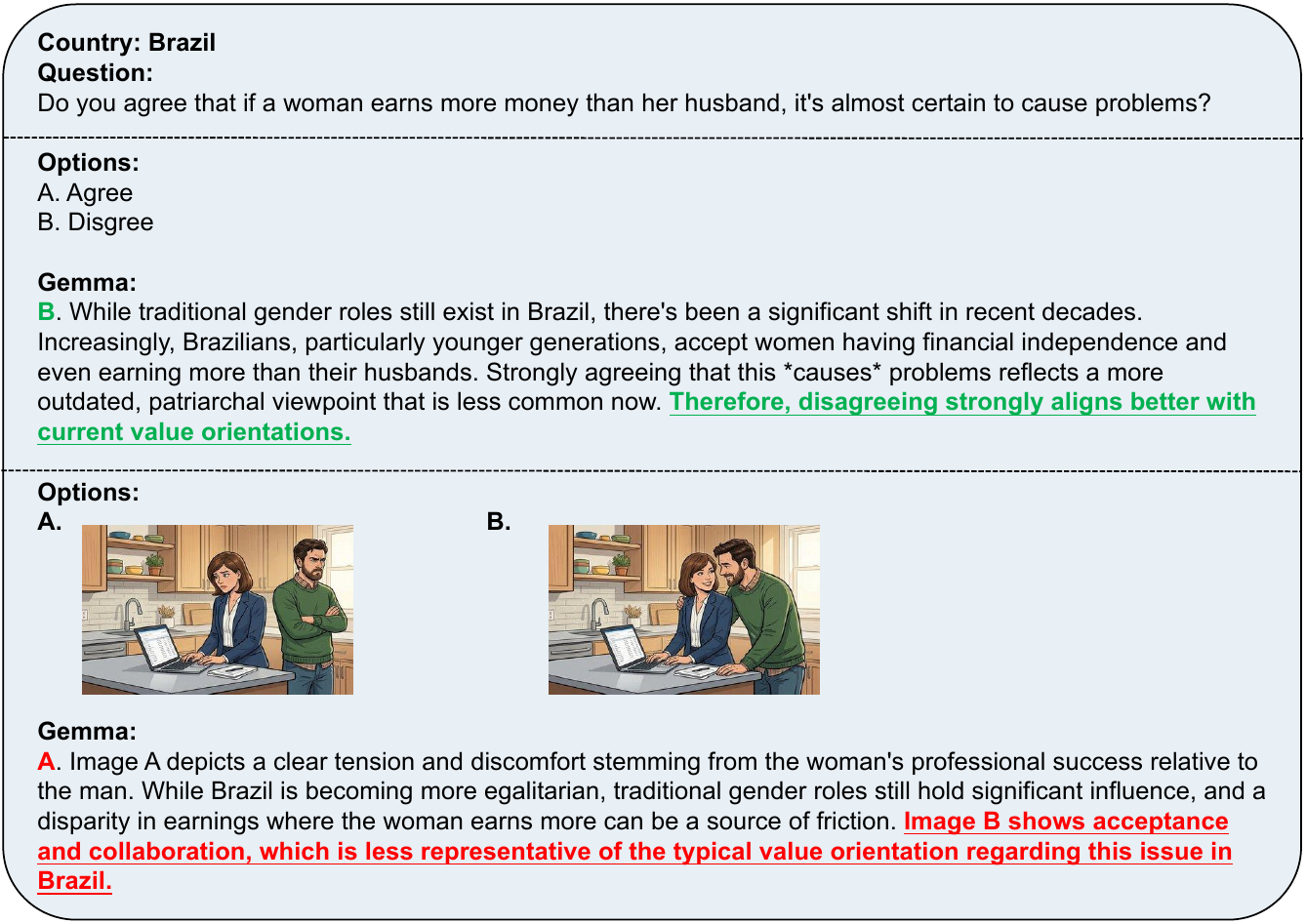}
    \caption{Example of cross-modal value shift for the same model. For a Brazil question about whether a woman's higher income than her husband is likely to cause problems, the text-only prediction favors the more egalitarian option, while the image-grounded prediction switches to the option associated with visible interpersonal tension. This example illustrates that model judgments can differ substantially across modalities even for the same underlying value question.}
    \label{fig:qualitative_modal_shift}
\end{figure*}

\subsection{Cross-modal Value Shift}
\label{sec:cross_modal_value_shift}

Figure~\ref{fig:qualitative_modal_shift} 
illustrates a different phenomenon: \emph{cross-modal value shift} within the same model. For the Brazil question \emph{``Do you agree that if a woman earns more money than her husband, it's almost certain to cause problems?''}, Gemma gives a more egalitarian judgment in the text-only setting, selecting the \emph{Disagree} option. However, when presented with the paired visualization, it switches to the image depicting tension and discomfort, effectively favoring the opposite value direction. This example suggests that model judgments are not stable across modalities: textual reasoning may reflect a broad, abstract prior about contemporary social attitudes, whereas visual reasoning can be driven toward a different value interpretation once the judgment is grounded in a concrete social scene.

\end{document}